\documentclass[journal]{IEEEtran}
\def \etal {\textit{et al.}}
\usepackage{gensymb}

\usepackage{subfigure}
\usepackage{color}
\usepackage{graphicx} 
\usepackage{amsfonts} 
\usepackage{amsmath}
\usepackage{siunitx} 
\usepackage{rotating}
\usepackage{cite}
\usepackage{enumitem}
\def \etal {et al.}

\ifCLASSINFOpdf
\else
\fi
\begin{document}
%
\title{The Multimodal Driver Monitoring Database: A Naturalistic Corpus to Study Driver Attention}
%
%
%

\author{Sumit Jha,~\IEEEmembership{Student~Member,~IEEE,}  Mohamed F. Marzban, \IEEEmembership{Student~Member,~IEEE,}  Tiancheng Hu, \IEEEmembership{Student~Member,~IEEE,} M.H Mahmoud, \IEEEmembership{Student~Member,~IEEE,} Naofal Al-Dhahir, ~\IEEEmembership{Fellow,~IEEE}, and Carlos~Busso, ~\IEEEmembership{Senior~Member,~IEEE,}
\thanks{S. Jha, M. Marzban, T. Hu, M. Mahmoud, N. Al-Dhahir, and C. Busso are with the Electrical and Computer Engineering Department, The University of Texas at Dallas, Richardson, TX 75080, USA. e-mail:(sumit.jha, mohamed.marzban, tiancheng.hu, mohamed.mahmoud, aldhahir, busso)@utdallas.edu.}
}

\maketitle

\begin{abstract}
A smart vehicle should be able to monitor the actions and behaviors of the human driver to provide critical warnings or intervene when necessary. Recent advancements in deep learning and computer vision have shown great promise in monitoring human behaviors and activities. While these algorithms work well in a controlled environment, naturalistic driving conditions add new challenges such as illumination variations, occlusions and extreme head poses. A vast amount of in-domain data is required to train models that provide high performance in predicting driving related tasks to effectively monitor driver actions and behaviors.  Toward building the required infrastructure, this paper presents the \emph{multimodal driver monitoring} (MDM) dataset, which was collected with 59 subjects that were recorded performing various tasks. We use the Fi-Cap device that continuously tracks the head movement of the driver using fiducial markers, providing frame-based annotations to train head pose algorithms in naturalistic driving conditions. We ask the driver to  look at predetermined gaze locations to obtain accurate correlation between the driver's facial image and  visual attention. We also collect data when the driver performs common secondary activities such as navigation using a smart phone and operating the in-car infotainment system. All of the driver's activities are recorded with high definition RGB cameras and time-of-flight depth camera. We also record the \emph{controller area network-bus} (CAN-Bus), extracting important information. These high quality recordings serve as the ideal resource to train various efficient algorithms for monitoring the driver, providing further advancements in the field of in-vehicle safety systems. 
\end{abstract}


%
\IEEEpeerreviewmaketitle




\section{Introduction}

\IEEEPARstart{T}{he} \emph{National Highway Traffic Safety Administration} (NHTSA) reported that in 2019, around 36,120 people died in car accidents across the United States \cite{NHTSA_2019}. Driver distractions and human errors are the leading causes of these accidents. With the growing use of in-vehicle technologies and hand-held devices, drivers are exposed to more distractions, leading to more accidents. Artificial intelligence-based driver behavior monitoring systems can play a key role in modeling and understanding driver distractions, triggering timely alarms to save humans lives. In addition, they can decide when to take over the vehicle's control from the driver in levels 3 and 4 of autonomous driving. These driver monitoring systems need to be trained with an annotated dataset that mimics practical driving environments. 

Recent advancements in the field of computer vision and deep learning have attracted increased research interests in the field of human-centered computing. The facial images can be used to extract multiple information such as identity \cite{Bronstein_2003,Parkhi_2015}, emotions \cite{DeSilva_1997}, and gaze \cite{Jha_2019,Li_2018}. While in theory these solutions should be directly applicable for developing in-vehicle safety systems, in practice, most of these systems fail to work in a driving environment due to additional challenges, such as occlusions and illumination changes \cite{Jha_2017_2}. Since most recent technologies based on \emph{deep learning} (DL) rely on in-domain data with ground truth annotations, naturalistic driving recordings collected from multiple drivers and appropriate labels are needed to advance in-vehicle safety systems. This requirement is particularly important if the focus is to analyze the driver's visual attention in realistic driving environments. Driver's gaze is one of the most crucial features for modeling and classifying the driver's distractions. However, estimating the driver's gaze in naturalistic driving environments is a challenging task because of the regular illumination changes and common occlusions due to eye glasses. To tackle these challenges, previous studies have utilized the driver's head pose to obtain a coarse gaze estimation, which is relatively easy to estimate \cite{Jha_2017,Jha_2018_2, vora_2018, Jha_2018}. Other studies have included eye features to capture fine gaze changes \cite{tawari_2014_3, vasli_2016}. Understanding and capturing the interplay between the head pose and the eye movements is essential for any driver behavior monitoring system and any solid gaze estimation algorithm \cite{Jha_2016}.

This study presents the \emph{multimodal driver monitoring} (MDM) database, which was specifically designed to study the driver's visual attention. It addresses some of the challenges of existing driver monitoring datasets, including their limited size and narrow scope focusing solely on either the driver's head pose or the driver's gaze zones. The MDM corpus is a new multimodal dataset that considers both head pose and gaze direction to facilitate learning this complex interplay between the head pose, eye movements and the resulting gaze. We achieve this goal by providing frame-based information about head pose, and spatial positions for target markers inside and outside the vehicle that were glanced at by the drivers. The corpus is multimodal, where we use multiple RGB cameras and depth videos to record the drivers. We also have a camera recording the road. We have collected 59 different gender-balanced subjects spanning different ethnic groups. Our dataset captures primary (e.g., mirror checking, lane keeping, left/right lane changes, left/right turns) and secondary activities (e.g., navigation using Google maps, operating the radio). In addition, we ask the drivers to look at predefined targets inside and outside the vehicle, providing temporal gaze information. The protocol includes multiple markers on the windshield, mirrors, dashboard, radio and gear shifter. It also includes looking at other vehicles, billboards, buildings and street signs. In addition, we collect continuous gaze annotations by asking the driver to look at a moving fiducial marker outside while the vehicle is parked. We design a systematic approach to obtain the driver's head pose per-frame annotation using the Fi-Cap helmet \cite{Jha_2018}. Fi-Cap is a helmet with multiple fiducial markers that are easy to detect. The helmet is worn on the back of the head avoiding facial occlusions when the driver is recorded with frontal cameras. We provide manual annotations using the ELAN toolkit \cite{Wittenburg_2006}, describing gaze information and the secondary activities conducted during the recordings.

This study provides an initial analysis of the MDM corpus, showing that our dataset covers a wide range of head poses in all three rotation axes (pitch, yaw and roll). The analysis reflects the head pose diversity obtained by including the primary and secondary activities in our protocol. The analysis also demonstrates that the MDM corpus represents real driving environments. Furthermore, we evaluate the accuracy of the head pose annotations provided by the Fi-Cap helmet as a function of the number of detected fiducial points. Moreover, we present some preliminary analysis on head pose estimation using depth cameras, demonstrating the potential of using point cloud data for vision-based tasks inside the vehicle. In summary, the main features of this paper are:

\begin{itemize}[leftmargin=0.3em]
\vspace{-0.2em}
\setlength{\itemindent}{1em}
\setlength{\itemsep}{0cm}%
\setlength{\parskip}{0cm}%
\item A multimodal database with drivers performing activities in naturalistic driving conditions
\item Continuous and discrete instances of driver's faces associated with available ground truth gaze 
\item Continuous tracking of the head pose of the driver with reliable ground truth using the Fi-Cap helmet
\item Multiple sensors to capture high quality RGB videos, point cloud data, CAN-Bus information and audio recordings. 
\end{itemize}

The rest of this paper is organized as follows. Section \ref{sec:relatedWork} describes previously published databases for in-vehicle driver monitoring systems. Section \ref{sec:Protocol} presents the data collection protocol, describing the annotation process of various activities and gaze events. Section \ref{sec:HeadFromFi-Cap} shows how we utilize the Fi-Cap helmet to annotate the driver's head pose on a frame by frame basis. It also analyzes the reliability in the frame-based head pose annotations as a function of the number of fiducial points. Section \ref{sec:analysis} analyzes the head pose and gaze angle distributions in our dataset. This analysis section also briefly describes a method to estimate the driver's head pose using the depth cameras. Section \ref{sec:conclusion} concludes the paper summarizing the main features of our corpus.

\section{Related Work}
\label{sec:relatedWork}

The research community has made several efforts to construct high-quality and diverse driving datasets, especially those focusing on road scenes. Examples include the BDD100K corpus \cite{Yu_2020}, Mapillary Vistas Dataset \cite{Neuhold_2017}, Waymo Open Dataset \cite{Pei_2020} and nuScenes dataset \cite{Caesar_2020}. However, the MDM dataset focuses on the drivers, not the road scenes. Therefore, we will review relevant datasets that have focused on head pose estimation (Sec. \ref{ssec:relatedheadpose}), and gaze estimation (Sec. \ref{ssec:relatedgaze}). This section also reviews recent efforts to collect long-term driving datasets (Sec. \ref{ssec:relatedlong}).

\subsection{Databases Focusing on Head Pose Estimation}
\label{ssec:relatedheadpose}

Databases focusing on head pose estimation aim to provide frame-based annotations for the driver's head pose \cite{Martin_2012,Tawari_2014,Schwarz_2017,Roth_2019,Borghi_2017,Selim_2020}, which has a wide range of applications including coarse gaze estimation, driver behavior modeling, and \emph{human-computer interaction} (HCI) for entertainment purposes. Table \ref{tab:hptable} summarizes the main databases created to estimate head pose of the driver.

\begin{table*}[t]
\caption{A comparison of existing driving databases focusing on head pose estimation.}
  \centering
    \begin{tabular}{p{6em}|p{10em}p{15em}p{9em}p{5em}p{8em}l}
          \hline
          {Dataset\newline{}Year} & {Recording \newline{} Conditions} & Sensors & {Annotation \newline{} provided} & {\# Subjects\newline{}M/F} & {Length of data\newline{}\# Frames}\\
          \hline
          \hline
          Lisa-P\cite{Tawari_2014}\newline{} 2012\newline{} & Day and night driving& RGB-face(640x480,30fps)\newline{}RGB-back(640x480,30fps)\newline{}Mo-Cap & Head orientation\newline{}7 facial landmarks & 14 & {1.85 hours \newline{}199,934}\\
          \hline
          CoHMEt\cite{Tawari_2014}\newline{} 2014\newline{} &Suburban street, highway & RGB-face(640x480,30fps)\newline{}RGB-2sides(640x480,30fps) \newline{}IMU & Head orientation & - & {0.28 hours \newline{}30231}\\
        \hline
        DriveAHead\cite{Schwarz_2017} \newline{} 2017  & Sunny, rainy and foggy\newline{} small town, highway and parking& depth(TOF) (512x424) \newline{} IR camera (512x424)\newline{}Mo-Cap & Head pose \newline{}binary occlusions\newline{} glasses, sunglasses, occlusion & 20\newline{} 16/4 & - \newline{} 1M frames\\
        \hline
        DD-Pose\cite{Roth_2019}  \newline{} 2019  & Parked, highway, cities\newline{}look at markers, make phone calls, read shop name, interact with pedestrians  & RGB-back\newline{}Depth(stereo)-face(2048x2048) \newline{}mo-cap\newline{}CAN-BUS\newline{} & head pose\newline{}occlusion label\newline{}steering wheel info\newline{}vehicle motion info & 27 \newline{} 21/6 & -\newline{}330k\\
        \hline
        Pandora \cite{Borghi_2017} \newline{} 2017  & Simulator\newline{}  Constrained and  unconstrained movement & RGB-face(1920x1080) \newline{} depth-face(512x424)\newline{}IMU\newline{} & Head orientation\newline{}Shoulder skeleton joints\newline{}Shoulder pose & 22\newline{}12/10 & -\newline{}250k\\
        \hline
        AutoPose \cite{Selim_2020} \newline{} 2020  & Simulator\newline{} Look at marker\newline{} Constrained and unconstrained movement & RGB-full body(1920x1080,30fps)\newline{}Depth-full body(512x424,30 fps)\newline{}Infrared-full body(512x424,30fps)\newline{}Infrared-face(512x424, 60fps)\newline{}mo-cap& Head pose\newline{}6 gaze zone\newline{}driver activity\newline{}glasses yes/no\newline{}glass type & 21\newline{}10/11 & 4.63 hours(IR) + 1.46 hours(Kinect)\newline{}1M(IR) + 316k(Kinect)\\
        \hline
        MDM corpus  \newline{} 2020  & Parked, driving\newline{}look at markers, read road landmarks,  change radio station, navigation on phone & RGB-face(1920x1080,60fps)\newline{}RGB-back(1920x1080,60fps)\newline{}RGB-mirror (1920x1080,60fps)\newline{}RGB-road(1920x1080,60fps)\newline{} depth(ToF)-face (171x224,45 fps)\newline{}grayscale-IR LED(171x224,45 fps)\newline{}Fi-Cap\newline{}Microphone array(5)\newline{}CAN-BUS & Head pose\newline{}21 gaze targets\newline{}vehicle info\newline{}audio & 59 \newline{} 32/27 &  50.23 hours\\
        \hline
        \end{tabular}
        \label{tab:hptable}%
\end{table*}

Martin \etal \cite{Martin_2012} presented the Lisa-P dataset, which is a naturalistic driving dataset collected on the road. The recordings included a motion capture system (marker and camera) at the back of the driver to track the head movement. The authors placed a camera on the dashboard to capture the driver's face. The corpus provides annotations for roll, yaw and pitch head rotation of the driver extracted with the motion capture system. Additionally, the corpus provides manual annotations for seven key facial features (eye corners, nose corners and nose tips) for some video sequences every 5 to 10 frames. This dataset contains 14 video sequences with an average of 14,281 frames per video, at 30 \emph{frames per second} (fps) (i.e., about 111 minutes of recordings in total). The resolution of the camera is 640 $\times$ 480.

The CoHMEt database \cite{Tawari_2014} offers naturalistic driving recordings from three RGB cameras. The first camera was mounted on the car frame between the front mirror and the windshield. The second camera was placed on the front windshield. The third camera was mounted near the rear view mirror. The cameras collected images with a resolution of 640 $\times$ 480 at 30 fps. The corpus included two \emph{inertial measurement units} (IMU), one on the driver's head and the other fixed on the car. The difference between the two IMU results was used as the label for the roll, yaw, and pitch head rotations. The IMUs are reset around every 10 seconds to avoid drifts in the measurements.  There are 30,231 frames in total in this dataset, which translates to roughly 17 minutes of recording. The dataset includes driving scenes on the streets and freeways near the University of California at San Diego. 

The DriveAHead \cite{Schwarz_2017} is a more recent driving dataset. The recordings include videos using a Kinect V2 sensor, which provides infrared and depth captures at a resolution of 512 $\times$ 424. A 3D motion captioning system was used to track the rotation and orientation of the driver's head. In addition, the authors provided binary annotation of three types of occlusions: hair, eye glasses and self-occlusion. The database consists of 21 video sequences from 20 unique subjects (4 males, 16 females), each performing parking maneuvers, driving on highways and driving in a small town. The data collection took place in sunny, rainy and foggy weather conditions. In total, this dataset has about 1 million frames.

The \emph{Daimler TU Delft Driver Head Pose} (DD-Pose) Benchmark \cite{Roth_2019} utilizes one high-resolution stereo camera (2,048 $\times$ 2,048) capturing the driver's face, and a wide-angle RGB camera recording from the backside of the driver.  The authors used a 3D motion capturing system and provide rotation and translation labels for the driver's head. The corpus also provides occlusion labels for each frame (none, partial and full occlusion).  In addition, the corpus provides steering wheel and vehicle motion information. This dataset includes both naturalistic driving as well as induced driving conditions.  Naturalistic driving conditions consist of driving recordings in highways and big German cities, with complex traffic scenarios. The drivers were asked to read names of the shops along the street while driving. During the induced driving conditions, the car is parked. The driver is instructed to look at a series of target points in the car. The DD-Pose dataset comprises of 27 subjects (21 males, 6 females) with a total of 330k frames.

Some of the databases for head pose estimation were recorded in simulators. The Pandora Database \cite{Borghi_2017} uses a Kinect One camera mounted on the simulated dashboard. 
The resolution of the RGB camera was set to 1,920 $\times$ 1,080, while the resolution for the depth camera was set to 512 $\times$ 424. In this dataset, drivers were asked to perform driving-like maneuvers such as looking at the rear-view mirror and holding the steering wheel. Subjects were first asked to rotate their heads along one axis at a time.  Then, they were instructed to freely move their heads. To create more variations in head pose and shoulder pose, subjects wore eye glasses, sun glasses, scarves, caps, and used smartphones. The corpus provides labels for the head and shoulder pose. An IMU on the back of the subject's head was used to record the ground truth for the head pose. In total, the dataset has 250k frames. Another dataset collected in a car simulator is the AutoPose corpus \cite{Selim_2020}. This corpus relies on one dashboard IR camera running at 60 fps, and a Kinect V2 camera (RGB, Depth, Infrared), mounted at the rear view mirror, running at 30 fps. There are 21 subjects in this dataset. The recording protocol included a series of tasks such as performing head rotations and looking at markers at various locations. Each task was performed in three conditions: without eye glasses, with eye glasses, and with sun glasses. Head rotations were performed with and without the subject wearing a scarf.  Due to interference of the two cameras, each subject performed the entire protocol twice, recording the session with either the IR camera or the Kinect V2 camera. Head pose labels are provided using an OptiTrack Motion Capture System. Gaze labels associated with the markers are also provided. No primary driving activities were performed in this dataset due to its simple setup consisting of just a seat and a steering wheel, without a computer monitor simulating the road scene. In total, there are 1,018,885 IR images and 316,497 images from Kinect.

\begin{table*}[t]
    \caption{A comparison of existing driving databases focusing on gaze estimation.}
  \centering
  \fontsize{8}{9}\selectfont
    \begin{tabular}{p{6em}|p{10em}p{15em}p{9em}p{5em}p{8em}l}
    \hline
          Dataset \newline{} Year  & {Recording \newline{} Conditions} & Sensors & {Annotation \newline{} provided} & {\# Subjects\newline{}M/F} & {Length of data\newline{}\# Frames}\\
    \hline\hline
    Fridman et al.\cite{Fridman_2016} \newline{}2016 & Highway\newline{}voice control interface, select phone number& RGB-face(800x600,30fps) & 6 gaze zone & 50    & 17.23 hours \newline{} 1.86M 
    \\
    \hline
    DAD\cite{Qiu_2019_2} \newline{}2019 & Naturalistic driving \newline{}suburban, urban and highway & RGB-road(1600x1200,30fps)\newline{}RGB-face(1600x1200,30fps)\newline{}CAN-BUS\newline{}Eye tracking glasses\newline{} Physiological sensors & 4-layer driver behavior representation & 4 & 250 hours
    \\
    \hline

    DR(eye)VE\cite{Palazzi_2019} \newline{}2019 & Day and night\newline{}different weather conditions\newline{}countryside, downtown and highway & RGB-head mounted(1280x720,30fps)\newline{} road(1920x1080,25fps) & Continuous gaze map\newline{}GPS,Vehicle speed,Course & 8\newline{}7/1 & {6 hours\newline{}0.56M} \\
    \hline
    DG-Unicamp\cite{Ribeiro_2019}\newline{}2019 & Parked and day, night driving\newline{}look at markers& RGB-face(320x240,30fps)\newline{}Depth(stereo)-face(320x240,30 fps)\newline{}Infrared-face(320x240,30 fps) & 18 gaze zone & 45\newline{} 35/10 & {12 hours \newline{}1M} \\
    \hline
    DGW\cite{Ghosh_2020}\newline{}2020 &Parked and day, night driving\newline{}look at markers& RGB-face & 9 gaze zone & 338\newline{} 247/91 & {N/A} \\
    \hline
    AutoPose\cite{Selim_2020} \newline{} 2020  & Simulator\newline{} Look at marker\newline{} Constrained and unconstrained movement & RGB-full body(1920x1080,30fps)\newline{}Depth-full body(512x424,30 fps)\newline{}Infrared-full body(512x424,30fps)\newline{}Infrared-face(512x424, 60fps)\newline{}mo-cap& Head pose\newline{}6 gaze zone\newline{}driver activity\newline{}glasses yes/no\newline{}glass type & 21\newline{}10/11 & 4.63 hours(IR) + 1.46 hours(Kinect)\newline{}1M(IR) + 316k(Kinect)\\
    \hline
    MDM corpus  \newline{} 2020  &Parked, driving\newline{}look at markers, read road landmarks,  change radio station, navigation on phone & RGB-face(1920x1080,60fps)\newline{}RGB-back(1920x1080,60fps)\newline{}RGB-mirror (1920x1080,60fps)\newline{}RGB-road(1920x1080,60fps)\newline{} depth(ToF)-face (171x224,45 fps)\newline{}grayscale-IR LED(171x224,45 fps)\newline{}Fi-Cap\newline{}Microphone array(5)\newline{}CAN-BUS & Head pose\newline{}21 gaze targets\newline{}vehicle info\newline{}audio & 60 \newline{} 32/27 &  50.23 hours\\
    \hline
    \end{tabular}%
  \label{tab:gazetable}%
\end{table*}%

Compared to the aforementioned driver head pose estimation dataset, our proposed MDM has several features that make this corpus an ideal platform to build machine learning algorithms to track head pose in a vehicle: (1) it includes data when the driver is operating the car and when the car is parked, (2) it provides complementary multimodal sensors (four different RGB, depth, grayscale-IR LED, microphone array, CAN-Bus), (3) it has frame-based head pose information obtained with a simple setting based on fiducial markers, and (4) it is recorded from 59 gender balanced subjects.

\subsection{Databases Focusing on Gaze Estimation}
\label{ssec:relatedgaze}

Driver gaze monitoring databases often provide ground truth location for regions where the drivers are asked to glance \cite{Fridman_2019,Palazzi_2019, Ribeiro_2019}, which can be used to train different gaze estimation algorithms. Tables \ref{tab:gazetable} summarizes the main databases created to estimate the gaze of the driver.

Fridman \etal \cite{Fridman_2016} constructs a naturalistic driving dataset. The subjects were asked to drive on a highway, and perform secondary tasks such as making phone calls and entering addresses into a navigation system using a voice-control feature. The driver's face was recorded using a resolution of 800 $\times$ 600 at 30 fps. The dataset includes recordings of 1,860,761 frames from 50 subjects (17h, 14m). The ground truth of the driver's gaze is manually labeled as one of six predefined gaze zones.

The DR(eye)VE Project \cite{Palazzi_2019} features data from both an ego-centric view of the road (from the eye-tracking glasses),  and a car-centric view (a camera capturing the road). This dataset is somehwat different from many other gaze datasets in that it provides gaze data from the ego-centric view. The eye-tracking is recorded at 60 fps, while the camera on the eye-tracking glasses acquired data at a resolution of 1280 $\times$ 720 at 30 fps. The road camera captured data at a resolution of 1920 $\times$ 1080 at 25 fps. In this dataset, eight subjects were asked to drive in various scenes including countryside, downtown and highway, during different hours during the day (both day and night) and in different weather conditions. This dataset provides ground truth gaze in the form of a continuous gaze map, derived from the eye-tracking glasses. In addition, the corpus provides \emph{global positioning system} (GPS) information and vehicle speed. In total, there are 555,000 frames in this dataset (roughly 6 hours).


Another interesting database is the \emph{driving anomaly dataset} (DAD) \cite{Qiu_2019_2}. The corpus was collected in a real car in an Asian city. The database involves two cameras, one capturing the road and the other capturing the driver's face. It also includes eye tracking glasses to obtain the gaze of the driver. Additionally, the vehicle information was collected from the CAN-Bus. The drivers were asked to use wearable devices, which captured \emph{heart rate} (HR), \emph{breath rate} (BR), and \emph{electrodermal activity} (EDA). One of the key strengths of this corpus is the detailed annotations, providing a 4-layer driver behavior representation: goal-oriented action (e.g., right turn), stimulus-driven action (e.g., stop), cause (e.g., a stopped car in front of ego car) and attention (e.g., pedestrian near ego lane). In total, the corpus has about 250 hours, where 120 hours have been annotated.

Ribeiro and Costa \cite{Ribeiro_2019} presented the DG-Unicamp corpus, which is a driving dataset that focuses on gaze zone estimation. It uses an Intel Realsense R200 camera placed on top of the dashboard. The camera provides aligned RGB and depth recordings as well as two sources (from stereo) of infrared recordings at 30 fps, with a resolution of 320 $\times$ 240. This corpus divided the gaze direction into 18 gaze zones, providing the corresponding gaze labels. The recordings were collected in a parked car due to safety concerns. The recordings were collected in different locations, light conditions (both day and night) at different hours to ensure maximum diversity in the recordings. Each subject was asked to look at each gaze zone for 10s and then move to the next gaze zone. The corpus consists of 45 subjects (35 males, 10 females), where five subjects were recorded twice.  In total, the corpus has about 1 million frames (12 hours). A similar corpus is the \emph{Driver Gaze in the Wild} (DGW) Dataset \cite{Ghosh_2020}, which  is a gaze zone classification dataset that features 338 subjects with age between 18 and 63 years old. In this dataset, subjects were asked to look at markers in different locations in a parked car at different locations in a university. The recording was collected at different times of the day and at night. During the recording, the driver was asked to fixate on one of the nine gaze markers and say its number. They directly labeled the gaze zone using an \emph{automatic speech recognition} (ASR) system.

We note that many gaze datasets are collected in parked conditions. We have observed that gaze behaviors are different from the patterns observed while the driver is operating a vehicle \cite{Jha_2016}. The driver is cognitively busy with primary driving tasks, resulting in more eye movements and shorter glances. Additionally, naturalistic driving conditions pose extra challenges due to sudden illumination changes and occlusion of the driver's face. These variations create a domain-shift problem for machine learning algorithms to be deployed in naturalistic driving. In the MDM corpus, we address these challenges by recording our database while the car is parked and while the driver is operating the vehicle. Likewise, compared to the aforementioned gaze estimation datasets, our proposed MDM corpus includes multiple modalities (four different RGB, depth, infrared, microphone array, CAN-BUS), provides fine-grained labels (21 gaze targets during driving condition; continuous label during parked condition).

\subsection{Large-Scale Naturalistic Databases for Visual Attention}
\label{ssec:relatedlong}

This section presents large-scale naturalistic databases that can be used to analyze driver visual attention. The NHTSA conducted a large-scale naturalistic driving study \cite{Neale_2005}, which involved 241 drivers with 100 cars equipped with various sensors, conducted over the course of 18 months. In this study, the subjects were asked to drive naturally in cars with sensors. These sensors included five cameras capturing the driver's face, driver side road view, passenger side road view, road, and over-the-shoulder view for the driver's hands and surrounding areas. The recordings also included the  CAN-Bus, a Doppler radar, an accelerometer, a gyroscope, and a GPS. The equipment also included a toolbox for drivers to identify accidents. NHTSA collected approximately 43,000 hours of data, which included 82 total collisions. In each accident, Neale \etal \cite{Neale_2005} identified the pre-event maneuver, precipitating factor, event type, contributing factors, associative factors and avoidance maneuver. 

The SHRP 2 Naturalistic Driving Study \cite{Campbell_2012} is the largest long-term naturalistic driving study. The goal of this study was to understand how drivers interact with and adapt to the environment, including the vehicle, the traffic environment, roadway characteristics, and traffic control devices.  It includes nearly 2,360 subjects from different parts of the U.S. The experimental vehicles had four cameras, recording at 15 fps, capturing the driver, road, instrument cluster, and road in the back. It had one camera capturing still images of passengers every few seconds. It also had information about the CAN-Bus, GPS, radar, accelerometer, and rate sensor information. The database also had a toolbox for drivers to identify accidents.

The MIT Advanced Vehicle Technology (AVT) Study \cite{Fridman_2019} is a long-term driver behavior monitoring dataset. Subjects are asked to drive either their own vehicles or MIT-provided vehicles for a period of time between one month and over a year. The cars are equipped with sensors. This study aimed at capturing all aspects of the drivers' states as well as their interactions with technologies such as autopilot. They have recordings of 3 or 4 Logitech C920 RGB cameras, facing the driver's face, body, road and occasionally the instrument cluster. The cameras record at 1,280 $\times$ 720 at 30 fps. In addition, they also have CAN-Bus data, audio from each of the cameras, IMU and GPS data. The data are synchronized via a customized board.

While these efforts provide attractive platforms to build driver behavior models for in-vehicle systems, they do not necessary focus on visual attention. Also, the data are proprietary and have important privacy constraints that prevent distributing the recordings to other research groups.

\begin{figure}[t]
\centering
\includegraphics[width=0.98\columnwidth]{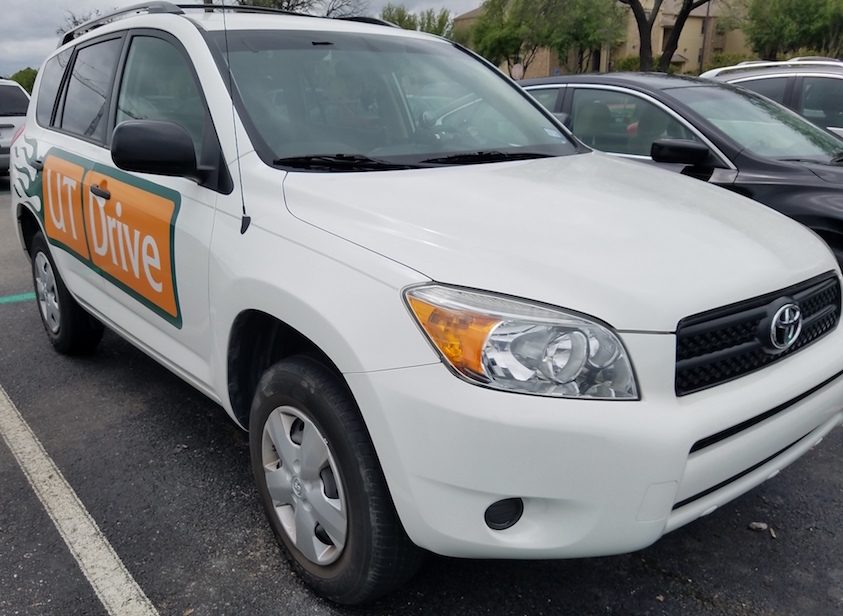}
\caption{The UTDrive vehicle used for the data collection. The car is equipped with multiple sensors.}
\label{fig:utdrive}
\end{figure}

\section{Data Recording Protocol}
\label{sec:Protocol}

Our goal is to create a multimodal database recorded from multiple drivers targeting applications that model driver visual attention. For this purpose, we collect the \emph{multimodal driver monitoring} (MDM) database, where our subjects are asked to follow a predefined protocol aiming to obtain labeled data to study gaze detection and head pose estimation. We perform our experiments in the UTDrive vehicle \cite{Hansen_2017,Angkititrakul_2007_2,Angkititrakul_2007}, which is a 2004 Toyota RAV4 SUV equipped with multiple on-board sensors and equipment (Fig. \ref{fig:utdrive}). We use various devices and sensors to set up baselines and record the data. The data is collected from a total of 59 subjects (32 females, 27 males). During the data collection, 31 of them wore prescription eye glasses, 2 wore sun glasses, and 26 did not wear any eye glasses. The drivers were all affiliated with the University of Texas at Dallas. We collected a total of 50 hours and 14 minutes of data, which captures the drivers performing a variety of activities as defined in our data collection protocol. We present in details the protocol followed during the data collection (Sec. \ref{ssec:protocol}), the sensors used to record the driver, car and road information (Sec. \ref{ssec:sensors}), the synchronization of the sensors (Sec. \ref{ssec:Synchronization}),  the methods used to spatially calibrate the orientation of different sensors in the environment (Sec. \ref{ssec:cam_calib}), and the annotations provided in the corpus (Sec. \ref{ssec:annotation}).

\subsection{Protocol}
\label{ssec:protocol}
Our protocol is an improved version of the data collection protocol used in our  previous work \cite{Jha_2016}, where we made several additions and improvements, learning from our previous experience. Our goal is to obtain (1) continuous gaze data with reliable ground truth, and (2) naturalistic data when the driver is performing driving related tasks. We ask the drivers to perform multiple tasks while being at the driver seat in the UTDrive car. We provide a detailed description of each of the tasks included in this protocol.

\subsubsection{Parked car: Looking at a target outside the vehicle}
\label{sssec:1}
\begin{figure}[t]
\centering
\includegraphics[width=0.98\columnwidth]{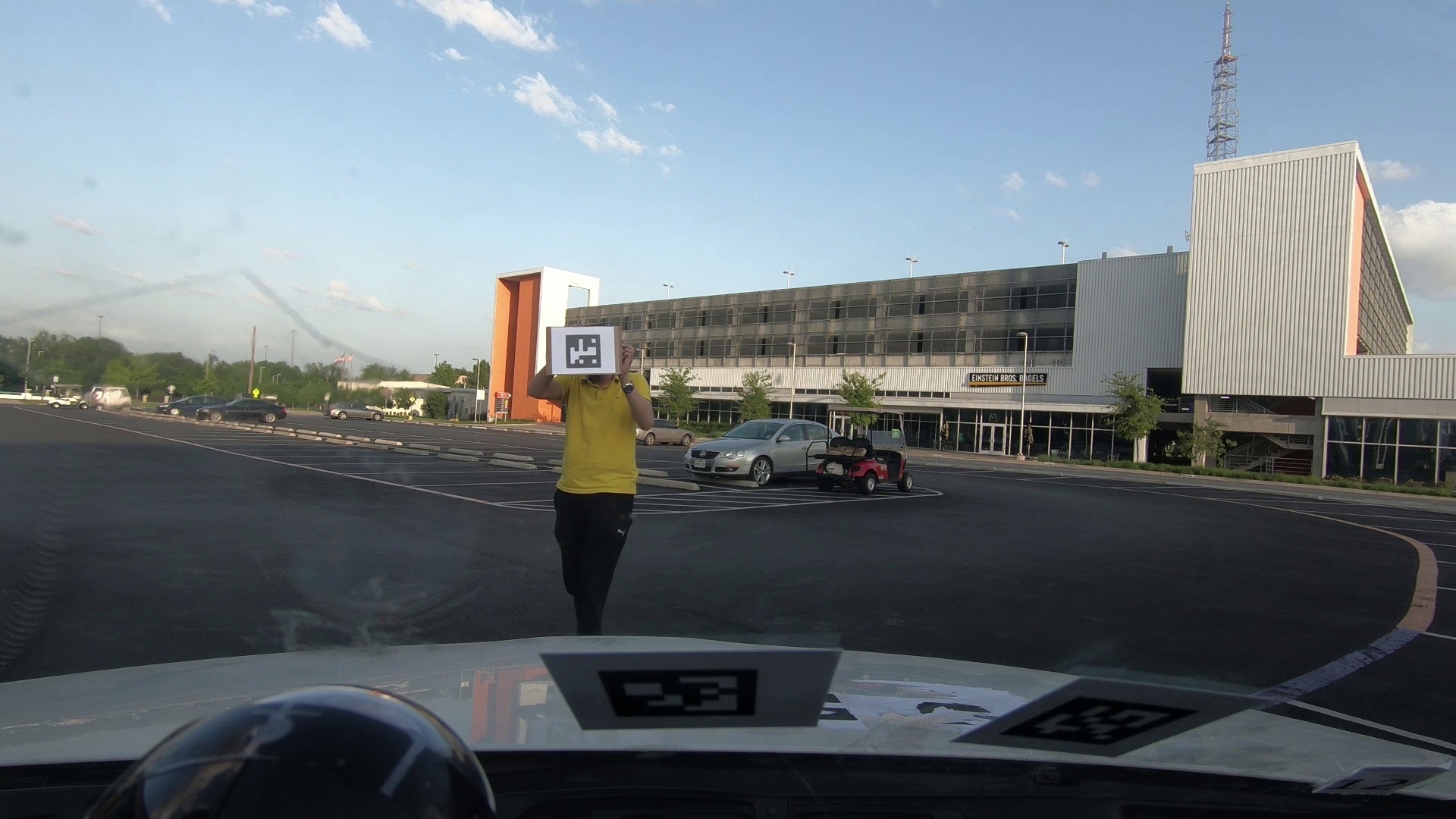}
\caption{Step 1 in the data collection protocol. The driver looks at a target board with a fiducial mark held  by one of the researchers outside the car. This part in the protocol provides continuous gaze information.}
\label{fig:atcont}
\end{figure}

The first step in our protocol aims to collect continuous gaze information. We start the recordings while the vehicle is parked. Then, we ask the driver to sit in the driver seat while a researcher slowly moves a board outside the vehicle (Fig. \ref{fig:atcont}). The board has a big fiducial marker with fixed pattern that can be easily tracked using basic image processing. We use an AprilTag \cite{Olson_2011} for this purpose such that we can easily track the 3D position of the marker (Fig. \ref{fig:atcont}). We ask the driver to follow the target board with her/his gaze as the researcher moves the board in front of the car. We collect data in three to five sessions with short breaks in between. Each session is approximately one minute long.

\subsubsection{Parked car: Looking at target markers inside the car}
\label{sssec:2}

\begin{figure}[t]
  \centerline{\includegraphics[width=0.98\columnwidth]{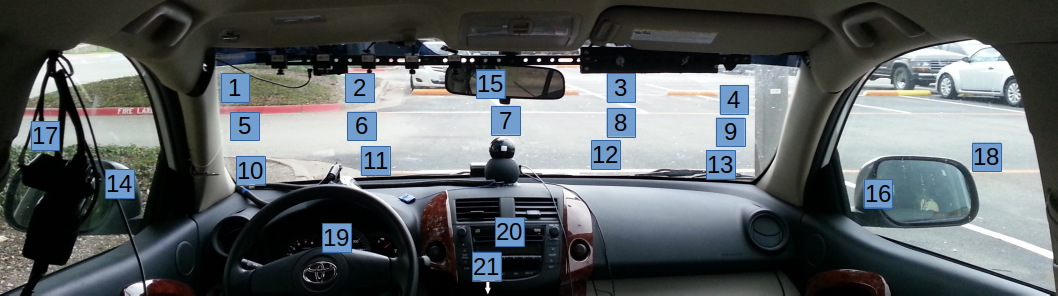}}
\caption{Location of the target markers placed inside the car. Markers 1-13 are in the windshield. The rest of the markers are in the mirrors (14-16), side of the windows (17-18), speedometer panel (19), radio (20), and gear (21).}
\label{Fig:marker}
\end{figure}

As mentioned before, our goal is to have gaze direction information. The second step in the protocol is to ask the driver to look at predefined markers placed inside the vehicle. We placed 21 different target markers inside the car  (Fig. \ref{Fig:marker}). The first 13 markers are placed on the windshield. The rest of the markers are placed on the mirrors (14-16), side of the windows (17-18), the speedometer panel (19), radio (20), and gear shifter (21). We ask the driver to look at each of these markers multiple times, calling the number in random order. The driver is asked to look at each marker as naturally as possible. This part of the protocol is conducted while the car is parked. The drivers get familiar with the location of the markers in a safe environment.

\subsubsection{Driving car: Looking at target markers inside the car}
\label{sssec:3}

After the subject is familiar with the task of looking at target markers, we repeat this step while the driver is operating the vehicle. We ensure this step is conducted when the subject is driving on a straight road, without making driving maneuver tasks. We ask the subject to quickly glance at the target marker. The investigator points to the location of the target markers to  reduce the cognitive load associated with searching for the right marker. 

\subsubsection{Driving car: Looking at targets outside the car}
\label{sssec:4}

We ask the subjects to locate and identify targets  such as landmarks, street signs, buildings and other cars on the road to complement the gaze data obtained while looking at the markers inside the car. We ask questions to the subjects that prompt them to naturally look at the target locations. Examples of these questions are: \emph{can you tell me the name of the store in the corner?} and \emph{can you read the license plate of the car in front?} We also ask the subjects to search for multiple targets (e.g., can you find red cars? and, can you read the license plates of as many cars as possible?). This part of the data collection creates implicit gaze responses that can be easily associated with the corresponding gaze directions.

\subsubsection{Driving the car: Navigation}
\label{sssec:5}
The next step of the data collection starts with the secondary tasks. We setup a multidestination navigation route on a smartphone. The directions are chosen to have multiple maneuvers. We ask the drivers to wait for the navigation instructions to reroute in situations when they miss an instruction. The audio of the navigation is turned off and the drivers has to follow navigation purely by looking at the phone screen. Once the drivers reach a destination, they have to hit the continue button to obtain the navigation instructions for the next stop in the trip.

\subsubsection{Driving car: Operating the radio}
\label{sssec:6}

The next step in the protocol is also a secondary driving task. We ask the drivers to operate the radio in the vehicle. We ask the drivers to turn on the car radio and set it to a given station. Since the UTDrive vehicle does not have radio controls on the steering wheel, the driver needs to operate the radio using the knobs in the central console. We repeat this step by calling out random stations for the drivers to find.

\subsubsection{Naturalistic driving}
\label{sssec:7}
During multiple instances in the data collection, we ask the subjects to drive with minimal intervention from the researcher. Only navigation instructions are provided to the driver and no additional task is carried out. These recordings aim to provide a baseline for natural driving conditions without secondary tasks.

\subsection{Multimodal Sensors}
\label{ssec:sensors}

\begin{figure}[t]
\centering
\includegraphics[width=0.98\columnwidth]{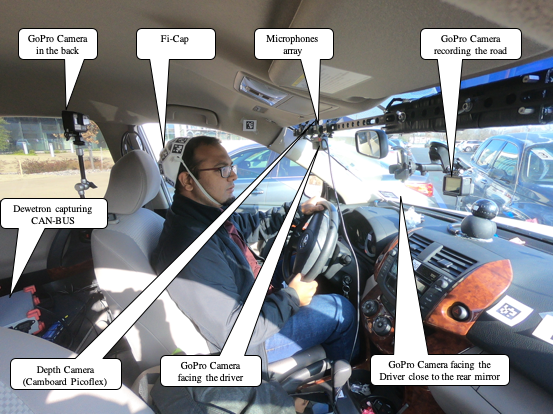}
\caption{Data collection setup showing the placement of the sensors in the UTDrive car. The setting includes four HD cameras, a depth camera, which simultaneously records grayscale images with IR illumination, microphone array and a Dewetron that records the CAN-Bus data.}
\label{fig:setup}
\end{figure}

To collect high quality data and relevant information from the driver, vehicle and road environment, we use multiple devices during the data collection.  Figure \ref{fig:setup} shows the setup used to place different devices inside the car. This section discusses the sensors used to collect the data.

\subsubsection{Fi-Cap}
\label{sssec:ficap}
\begin{figure}[t]
\centering
\includegraphics[width=0.75\columnwidth]{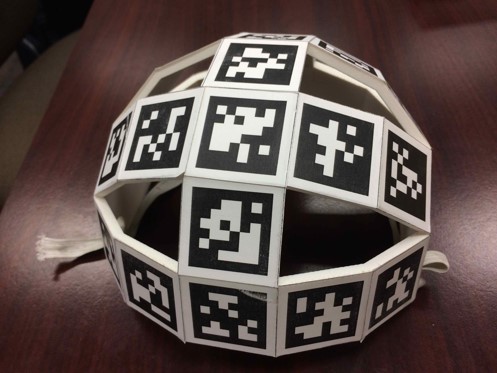}
\caption{The Fi-Cap helmet \cite{Jha_2018}, which we use to obtain head pose labels for each frame. A HD camera is placed behind the driver to record the fiducial points in the helmet (Fig. \ref{fig:bc}).}
\label{fig:ficap}
\end{figure}

Providing reliable labels for head pose estimation is important for modeling the driver's visual attention. Our primary objective was to have a label for each of the frames in our corpus. Alternative systems to track head pose in actual vehicle include motion capture systems, which   involves a complex setting \cite{Martin_2012}, and IMU measures, which are sensitive to drifts in the measurement. Our solution for estimating the frame-based labels for the head pose is based on fiducial markers placed on a solid helmet with a predefined structure with highly contrasting black and white patterns. The position and orientation of multiple fiducial points are easily tracked with simple algorithms, which are then used to estimate the head pose of the drivers. Our first prototype was a headband with 17 fiducial markers used in the forehead \cite{Jha_2016}. The main problem of this setting was the facial occlusion caused while wearing the headband. Our second prototype is the Fi-Cap helmet \cite{Jha_2018} (Fig. \ref{fig:ficap}), which is used in this data collection. The Fi-Cap is a helmet that contains 23 different AprilTags \cite{Olson_2011}. The size of each of them is 3.2cm $\times$ 3.2cm.  The high number of fiducial points, the size of each marker, and the 3D structure of the helmet lead to robust estimation of head movement, which can be established for a wide range of head poses. Another advantage of the Fi-Cap device is that the helmet is worn on the back of the driver's head, so the facial occlusion is minimal (see Fig. \ref{fig:blend}). We request the drivers to avoid touching and adjusting the Fi-Cap helmet during the recordings to avoid drifts. However, these drifts can be compensated over time during the calibration. The readers are referred to Jha and Busso \cite{Jha_2018} for more details about the Fi-Cap helmet. Notice that with the addition of the Fi-Cap helmet, we need an additional camera that can record the back of the driver's head. The additional camera is the only extra sensors in our setting, which makes this solution a very convenient approach for estimating frame-based labels for head poses.

\subsubsection{RGB Cameras}
\label{sssec:rgb}

\begin{figure}[t]
\centering
\subfigure[Face camera]
{
\includegraphics[width=4.cm]{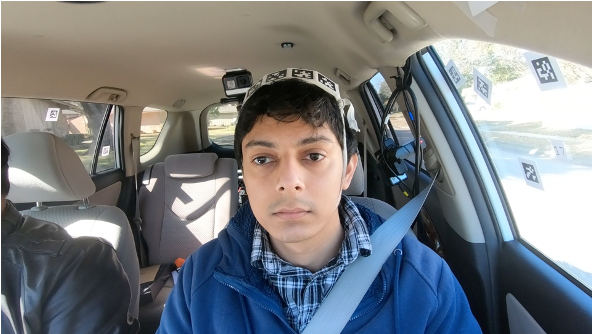}
\label{fig:fc}
}
\subfigure[Road camera]
{
\includegraphics[width=4.cm]{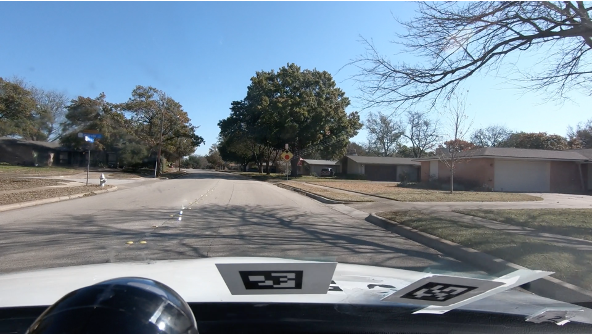}
\label{fig:rc}
}\\
\subfigure[Mirror camera]
{
\includegraphics[width=4.cm]{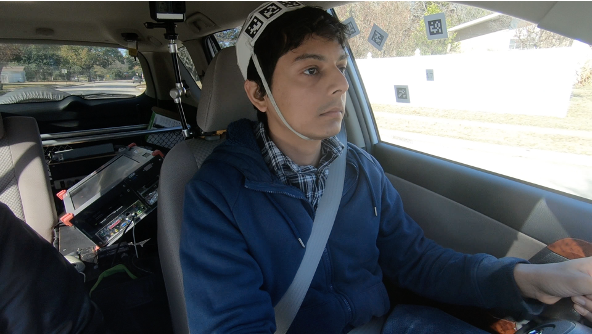}
\label{fig:mc}
}
\subfigure[Back camera]
{
\includegraphics[width=4.cm]{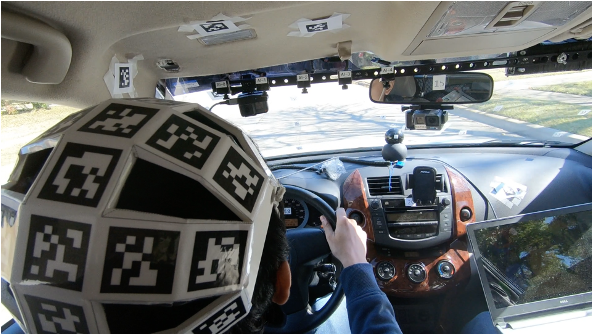}
\label{fig:bc}
} \\\hspace{-0.25cm}
\subfigure[Depth Camera]
{
\includegraphics[width=4.cm]{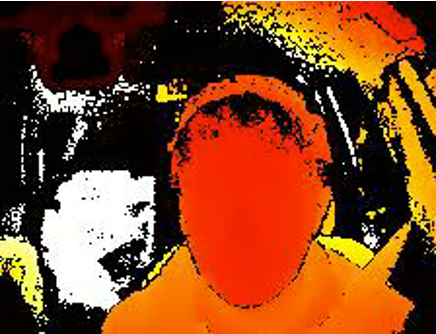}
\label{fig:pd}
}
\subfigure[Grayscale image with IR LEDs]
{
\includegraphics[width=4.cm]{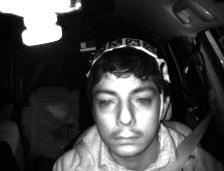}
\label{fig:ToF}
}
\caption{Examples of images collected with different sensors. All the recordings are synchronized with a clapboard.}
\label{fig:blend}
\end{figure}

We use four GoPro Hero 6 Black cameras in our data collection. All the cameras are set to record in full HD ($1,920 \times 1,080$) in linear mode. The purpose and placement of each of the four cameras are listed as follows:

\begin{itemize}[leftmargin=0.3em]
\vspace{-0.2em}
\setlength{\itemindent}{1em}
\setlength{\itemsep}{0cm}%
\setlength{\parskip}{0cm}%
\item Face Camera - This camera is placed in front of the driver to record the frontal view of the driver's face (Fig. \ref{fig:fc}). 
\item Road Camera - This camera is placed in the center of the dashboard facing the road. This camera captures road related information (Figure \ref{fig:rc})
\item Mirror Camera - This camera is placed under the rearview mirror to obtain a profile view of the driver's face. Several cameras installed on commercial cars are placed on the rearview mirror, so this camera provides realistic views that can be expect in deploying in-vehicle technology (Figure \ref{fig:mc})
\item Back Camera - This camera is placed behind the driver to record the Fi-Cap helmet (Figure \ref{fig:bc}). The camera is attached to the driver seat. 
\end{itemize}

\subsubsection{Depth Camera}
\label{sssec:depth}

A sensor that can be particularly useful in a vehicle is a depth camera, which has gained popularity in various computer vision applications. For our database, we place a camboard picoflex camera close to the face camera. This camera records the point cloud data using time-of-flight technology, providing robust estimation of the depth map in varying illumination conditions (Fig. \ref{fig:pd}). This camera provides reliable information even during night-time. The camera also records grayscale images which are illuminated using IR LEDs (Fig. \ref{fig:ToF}) and, hence, is immune to ambient lighting variations. The camboard picoflex camera records at 45fps, with a resolution of 224 $\times$ 171.

\subsubsection{CAN-Bus}
\label{sssec:canbus}

The UTDrive vehicle records the CAN-Bus information during the recordings. From the CAN-Bus, we obtain the information about accelerator, brakes, steering and speed of the vehicle. The UTDrive vehicle also has a gas and brake pressure sensors. The information provided in the CAN-Bus is very useful, for example, to analyze the vehicle state, and obtain driver maneuver information.

\subsubsection{Microphone Array}
\label{sssec:microphone}

The UTDrive vehicle is also equipped with a microphone array with five microphones. While our protocol does not include any task that elicits speech, the audio information is useful for understanding potential auditory distractions. 

\subsubsection{Dewetron}
\label{sssec:Dewetron}

The CAN-Bus data and the microphone array are connected to a Dewetron system, which stores and synchronizes the modalities.

\subsection{Synchronization of Sensors}
\label{ssec:Synchronization}

The Dewetron system can record up to two cameras. Since we are using multiple sensors, each with its own independent clock, we use a clapboard to synchronize the recordings. This is a simple, but effective approach. The clapboard provides a reference time to synchronize these devices. We perform two claps to synchronize all the cameras. The first clap is performed inside the car such that it is visible to the face camera, mirror camera, back camera and the depth camera. The second clap is performed outside the car such that it is visible to the back camera and the road camera. The claps can be precisely detected in the recording using the video and the audio. We did not observe any drift in the recordings provided by the GoPro cameras. The sampling rate for the picoflexx camera is not constant. However, the camboard picoflex camera records the time information of each frame, which is used to synchronize the point cloud with RGB data.

\subsection{Cameras Calibration}
\label{ssec:cam_calib}

We perform the camera calibration in two steps. In the first step, we establish a reference position and orientation for each camera. This step was performed once before we started our data collection. For this step, we place all the cameras at their desired locations. We place multiple AprilTags \cite{Olson_2011} of size $3.2 \rm{cm} \times 3.2 \rm{cm}$ around the car such that all the cameras can see some of the tags. Then, we use an additional camera that is used to capture static images inside the vehicle from multiple perspectives. The procedure is similar to the one proposed in Jha and Busso \cite{Jha_202x}, which relies on the Kabsch algorithm \cite{Kabsch_1978} to estimate transformation between coordinate systems. We establish the location of the tags from multiple cameras, which are used to obtain the relative position between the cameras.

\begin{figure}
    \centering
    \normalcolor
  \includegraphics[width=0.9\columnwidth]{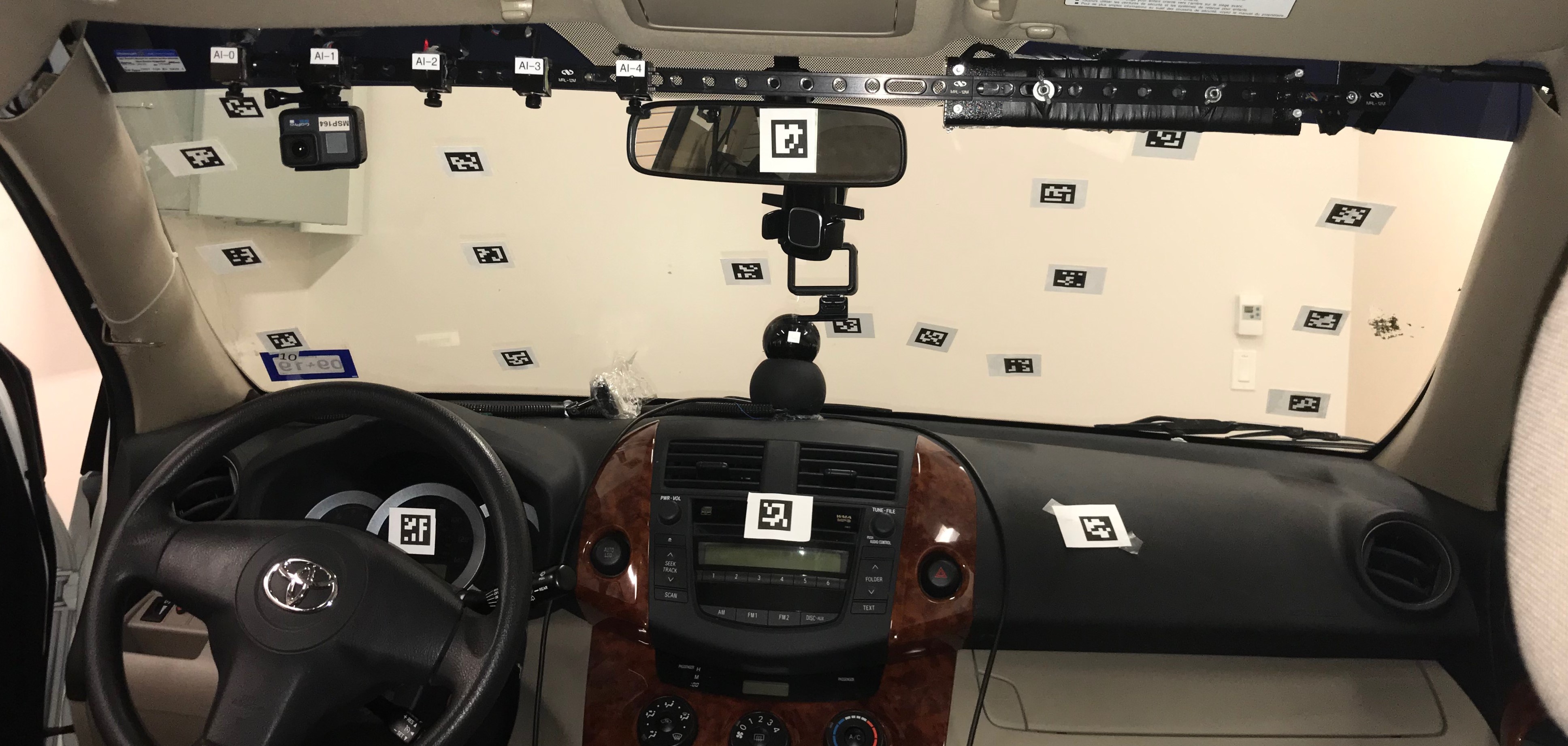}
  \caption{We place AprilTag on top of the target markers to estimate their exact position in the 3D space during the calibration.}
  \label{fig:ApriltagOnMarkers}
\end{figure}

We also place AprilTags at each of the target marker (Fig. \ref{Fig:marker}) as shown in Figure \ref{fig:ApriltagOnMarkers} to estimate the 3D positions of the markers. Using these tags, we obtain the 3D coordinates of the markers with respect to a common coordinate system (we arbitrarily choose the back camera as our reference).

The second step in the calibration process is to compensate for small camera placement variations between sessions. The driver seat is adjusted affecting the location of the back camera. In addition, all the cameras are removed from the vehicle after the session. After recharging the equipment, they are reinstalled back in the vehicle to their original positions. This process can result in slight changes in the positions and orientations of the cameras. For this purpose, we fix multiple AprilTag markers in different locations in the car, including the ceiling, windshield, dashboard and side-windows. These markers are left in the car for all the sessions. We ensure that each camera has at least two AprilTag markers within its field of view. By estimating the corners and the center of each AprilTag, we estimate at least $10$ reference points for every camera view. We calculate the variations in the setup of the cameras for each session using the changes in the positions of these tags from the original reference setup using the Kabsch algorithm.

\subsection{Annotations}
\label{ssec:annotation}

\begin{figure*}[t]
\centering
\includegraphics[trim=2.2cm 0 0 1cm, clip, width=14cm]{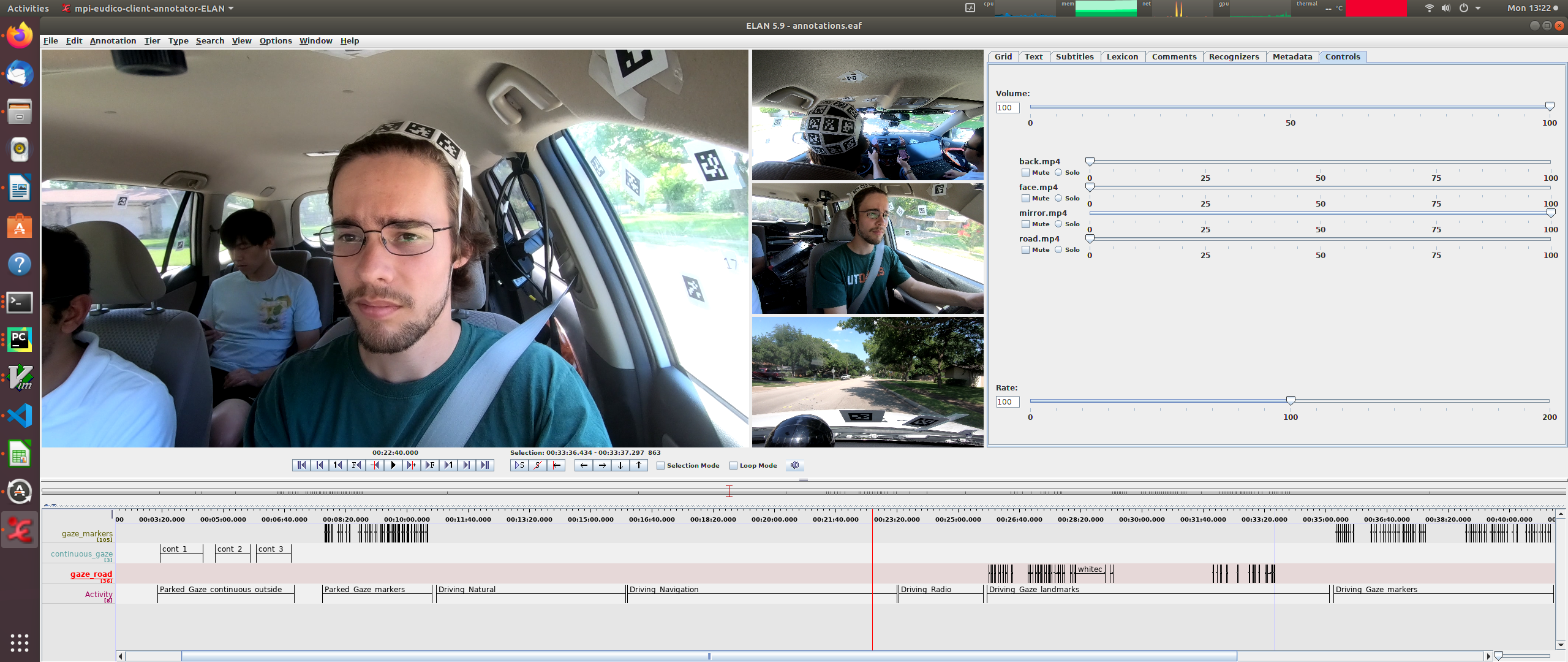}
\caption{The corpus uses ELAN as the interface to annotate the recordings. The annotation includes the segments associated with the seven steps in the protocol. It also provides the temporal information for instances when the subject looked at the target markers or objects inside and outside the vehicle.}
\label{fig:ELAN}
\end{figure*}

We use the ELAN tool \cite{Wittenburg_2006} to annotate events in the videos (Fig. \ref{fig:ELAN}). The videos from the four GoPro cameras are synchronized by manually finding the precise timings of the claps in each video. The annotation process has multiple channels with relevant information that are currently provided in the database:

\begin{itemize}[leftmargin=0.3em]
\vspace{-0.2em}
\setlength{\itemindent}{1em}
\setlength{\itemsep}{0cm}%
\setlength{\parskip}{0cm}%
\item Activity: This channel splits the data collection into the stages defined in our protocol (Section \ref{sec:Protocol}).
\item Continuous\_gaze: This channel identifies the segments when the subjects are following the target board outside the car while the car is parked (Section \ref{sssec:1}).
\item Gaze\_markers: This channel locates the actual times when subjects are looking at different markers inside the vehicle (Sections \ref{sssec:2} and \ref{sssec:3}). The numbers of the markers are also provided in the annotations. 
\item Gaze\_road: This channel locates the times when subjects are looking at different landmarks on the road (Sec. \ref{sssec:4}). It also provides information about the actual target objects glanced at by the driver. 
\end{itemize}

Since ELAN provides a convenient way of adding multiple tier of annotations, more information can be added in the future based on specific target applications (e.g., mirror-checking actions \cite{Li_2016}, driver maneuvers \cite{Hulnhagen_2010}).

\section{Head Pose Estimation with Fi-Cap}
\label{sec:HeadFromFi-Cap}

\subsection{Head Pose Estimation}
\label{ssec:EstimationFi-Cap}

The Fi-Cap contains multiple AprilTags that are tracked using the back camera. The locations of the fiducial points determine the frame-based labels for the head poses of the drivers. The first step to track the head pose is to obtain the position and the orientation of the cap with respect to a standard reference given a visible subset of tags on the cap. We follow the steps proposed in Jha and Busso \cite{Jha_2018}. We obtain a mesh of the Fi-Cap by finding the corners of each of the AprilTags in the structure by conducting continuous recording of the cap from different perspectives. Using this mesh as a reference, we obtain the transformation of all the visible corners in a given frame using the Kabsch algorithm \cite{Kabsch_1978}. We use a method inspired by the \emph{iterative closest point} (ICP) algorithm \cite{Arun_1987} to reduce the effect of outliers in estimating the pose. We reconstruct the mesh from the estimated transformation and remove the points that have high reconstruction error. Then, we recalculate the transformation matrix based on the reduced set of points. We repeat this step until the error is below a threshold. We consider this final transformation as the pose of the Fi-Cap helmet for each frame.

The next step is to find the head orientation from the Fi-Cap pose. We use the multiple local reference frame calibration approach proposed by Hu \etal \cite{Hu_202x}. This transformation is subject-specific because each subject might wear the Fi-Cap helmet differently. This transformation ensures that the ground truth for the head pose is defined uniformly across subjects. This approach uses the ICP algorithm \cite{Arun_1987} to achieve this goal. First, we manually select one frontal frame from one subject and use it as the \emph{global reference} (GR) frame. We calculate its head pose estimate using OpenFace2.0 \cite{Baltrusaitis_2018} by processing its corresponding gray-scale image. For every subject, we choose a number of frames with the closest rotation angles as the GR frame. We refer to these selected frames as \emph{local reference} (LR) frames. For each LR frame, we crop the face region in the point cloud data for the LR and GR frames. Then, we run the ICP algorithm between the two point cloud data to find a transformation between them, which compensates for the differences in head pose rotation between the LR and GR frames. Therefore, the LR and GR frames do not need to have the exact head rotation.  We maintain multiple transformations for each subject to ensure maximum accuracy against any movement of the Fi-Cap helmet during the recordings. 

\subsection{Reliability of the Fi-Cap Helmet}
\label{ssec:Ficapresult}

The Fi-Cap helmet provides the position and orientation of the driver's head. In Jha and Busso \cite{Jha_2018}, we estimate the reliability of the Fi-Cap using a virtual animation, where an exact copy of the helmet was rendered and placed in a virtual agent. By controlling the rotation, resolution of the image and illumination, we demonstrated that the median angular error was less $\ang{1.62}$, and that the 95 percentile error was less than $\ang{2.88}$. We also validated the approach in a laboratory setting, using a laser mounted on glasses worn by a subject. We projected the laser into a white screen, measuring the angular distance between the head pose direction, as determined by the laser marker, and our prediction. The results showed median angular errors that were less than 2.31\degree, and 95 percentile angular errors that were less than 6.93\degree. The estimation of head position is more reliable when more fiducial markers are detected. This section evaluates the reliability of the ground truth for head pose provided by the Fi-Cap helmet as a function of the number of detected fiducial markers. 

Our approach to evaluate the reliability of the Fi-Cap helmet consists of comparing the rotation matrix estimated with all the fiducial markers, and with a subset of the markers. Let $n$ denotes the number of tags detected in one particular frame. We first calculate the global head pose using all the $n$ tags. We refer to this rotation matrix as $R_\mathit{all}$. For every combination, we evaluate the rotation matrix with different subsets of markers using only the selected tags: 

\begin{equation}
\binom{n}{1}, \binom{n}{2}, \binom{n}{3}...\binom{n}{n-2}
\label{eq:binom}
\end{equation}

We refer to one of these matrices as $R_\mathit{subset}$. Then, we calculate the rotation difference between $R_\mathit{all}$ and $R_\mathit{subset}$, namely $R=R_\mathit{subset}R_\mathit{all}^{-1}$.  We aggregate all the matrices calculated in the previous step, obtaining their mean absolute value and standard deviation. Notice that we use the combinations up to $n-2$ to avoid having few combinations to estimate these statistics (Eq. \ref{eq:binom}). 

\begin{figure}[t]
\centering
\subfigure[Mean of the rotation angle of matrix $R$]{
\includegraphics[trim=5cm 5cm 5cm 8cm, clip, width=.95\columnwidth]{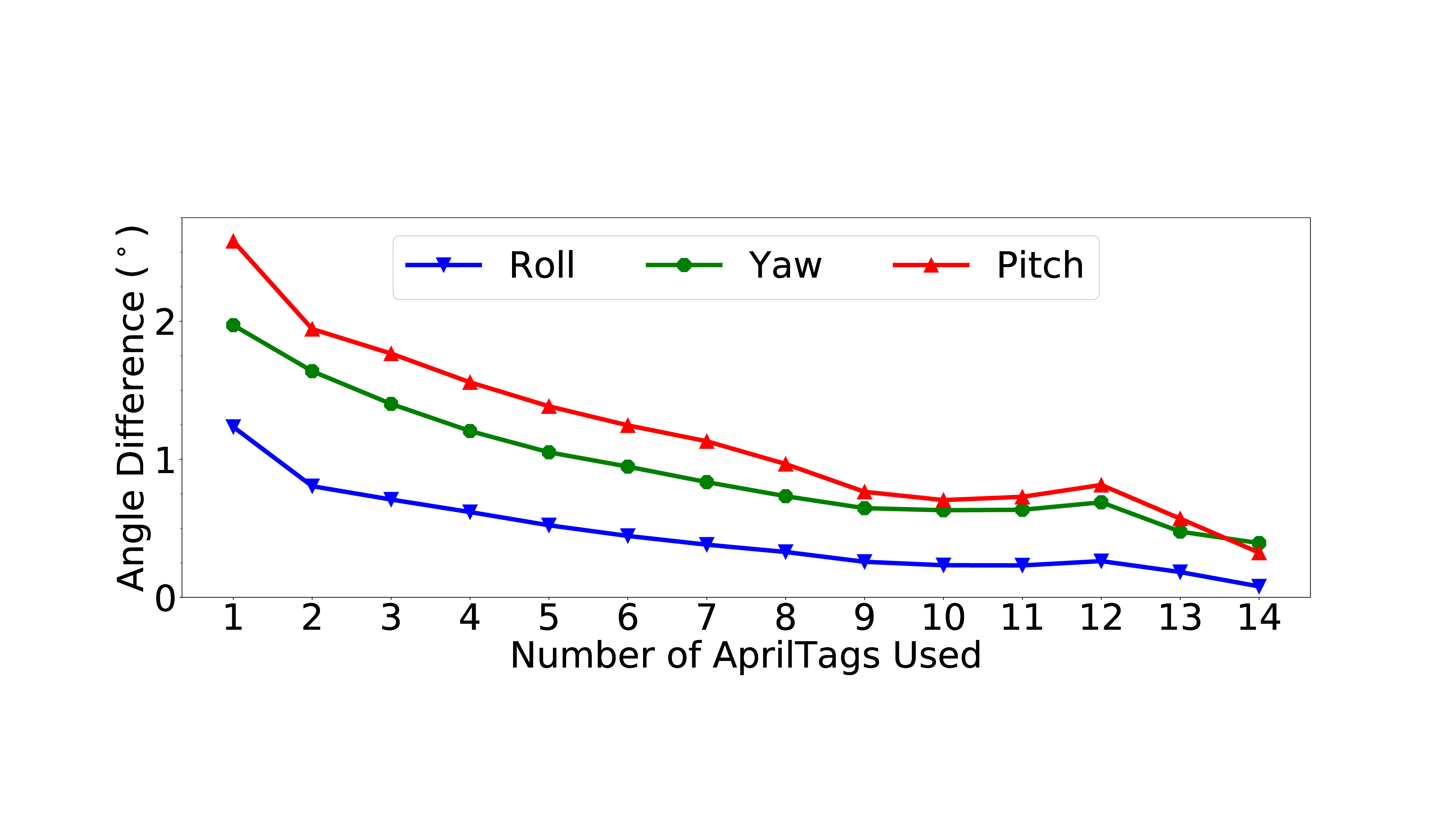}
}
\subfigure[Standard deviation of the rotation angle of matrix $R$]{
\includegraphics[trim=5cm 5cm 5cm 8cm, clip, width=.95\columnwidth]{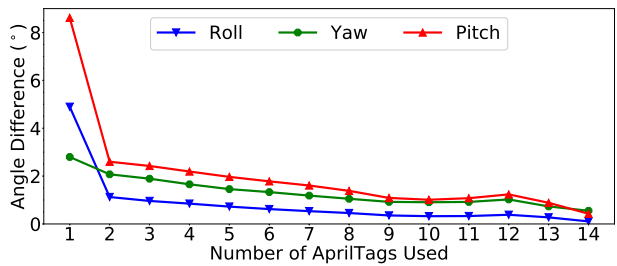}
}
\caption{Mean and standard deviation of the rotation angle of matrix $R$ capturing the variability of head rotation angles estimated as a function of the number of detected fiducial points.}
\label{fig:ficap_stability}
\end{figure}

Figure \ref{fig:ficap_stability} shows the mean and standard deviation of the rotation angle of matrix $R$ as a function of the number of fiducial points used in the estimation. Lower angles indicate small deviations from the angles estimated with all the markers. This analysis shows the uncertainty in the ground truth labels when few fiducial points are used in the estimation. We observe that we can obtain a mean difference of less than $\ang{1}$ with sufficient number of visible tags (2 for roll, 5 for yaw and 8 for pitch). We observe the largest difference for pitch angle, which is the most difficult rotation to estimate. The standard deviation represents the variability in estimating the Fi-Cap pose when selecting different subsets of visible tags. The standard deviation drops significantly when more than two tags are detected. As we add more fiducial points, the estimation of the Fi-Cap position becomes more consistent. In fact, Figure \ref{fig:ficap_stability} shows that with more tags, the rotation estimates are closer to $R_\mathit{all}$. As a reference, the average number of detected fiducial points per frame is 6.16. We have five or more fiducial points in 57.93\% of the frames with detected tags. 
These results show that our ground truth labels for head pose are very close to the hypothetical scenario where every single tag is visible and detected. This analysis demonstrates the reliability of our ground truth labels estimated with the Fi-Cap helmet. 

\section{Analysis}
\label{sec:analysis}
The MDM database provides a diverse resource to design algorithms to model the visual attention of the driver. This section analyzes the database to discuss its potential use. First, we study the distribution of head pose across different ranges of angles (Sec. \ref{ssec:DistributionHead}). Then, we study the distribution of the gaze data for different parts of the corpus (Sec. \ref{ssec:DistributionGaze}). Finally, we discuss a potential application of our corpus to predict the head pose of the driver using  point cloud data (Sec. \ref{ssec:depth}).

\subsection{Head Pose Distribution}
\label{ssec:DistributionHead}

\begin{figure*}[t]
\centering
\subfigure[Driver's head: yaw-pitch   angles]
{
\includegraphics[trim=0 0 2.7cm 0, clip, height=5.0cm]{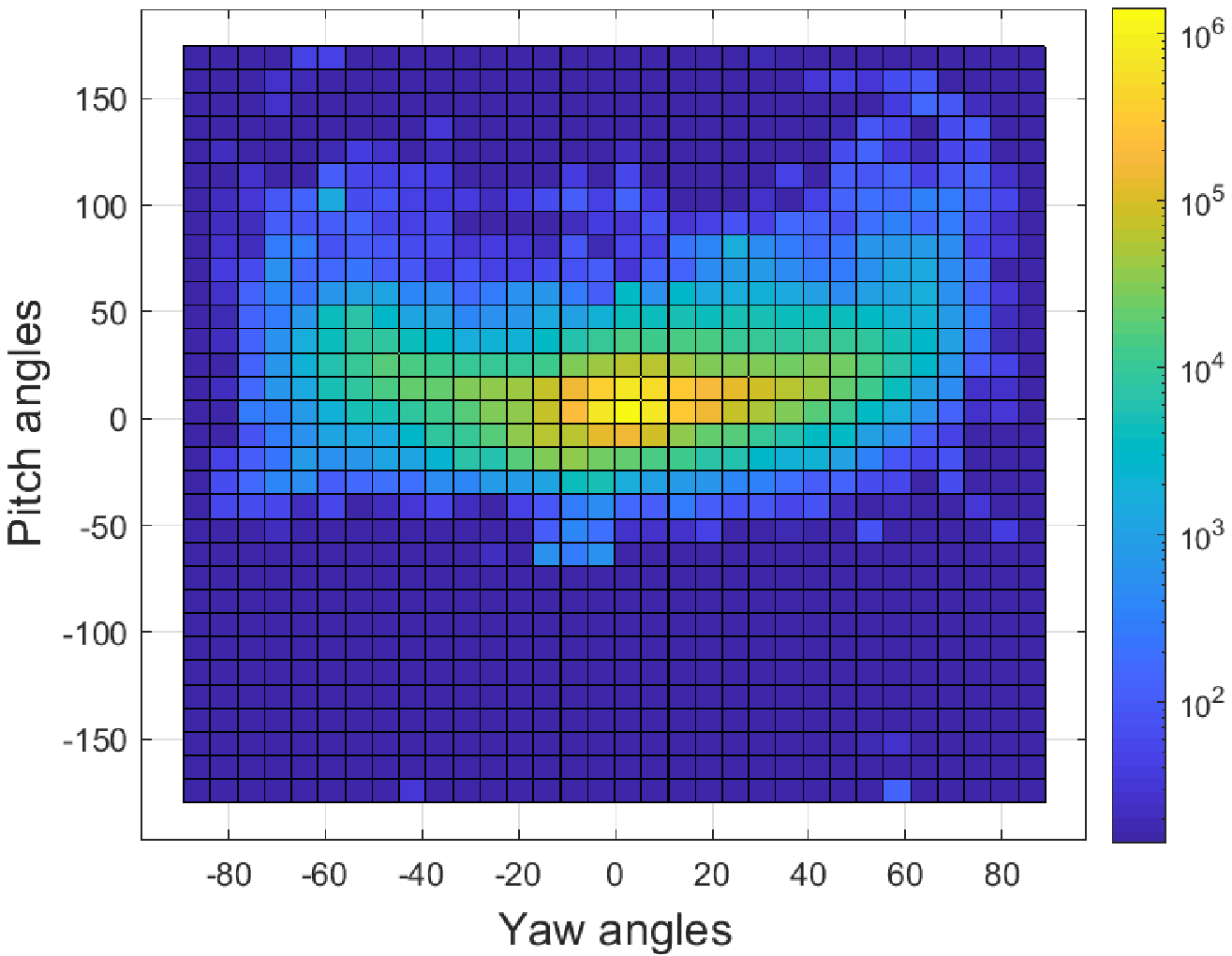}
\label{fig:yp}
}  \hspace{-0.5cm}
\subfigure[Driver's head: yaw-roll angles]
{
\includegraphics[trim=0 0 2.7cm 0, clip, height=5.0cm]{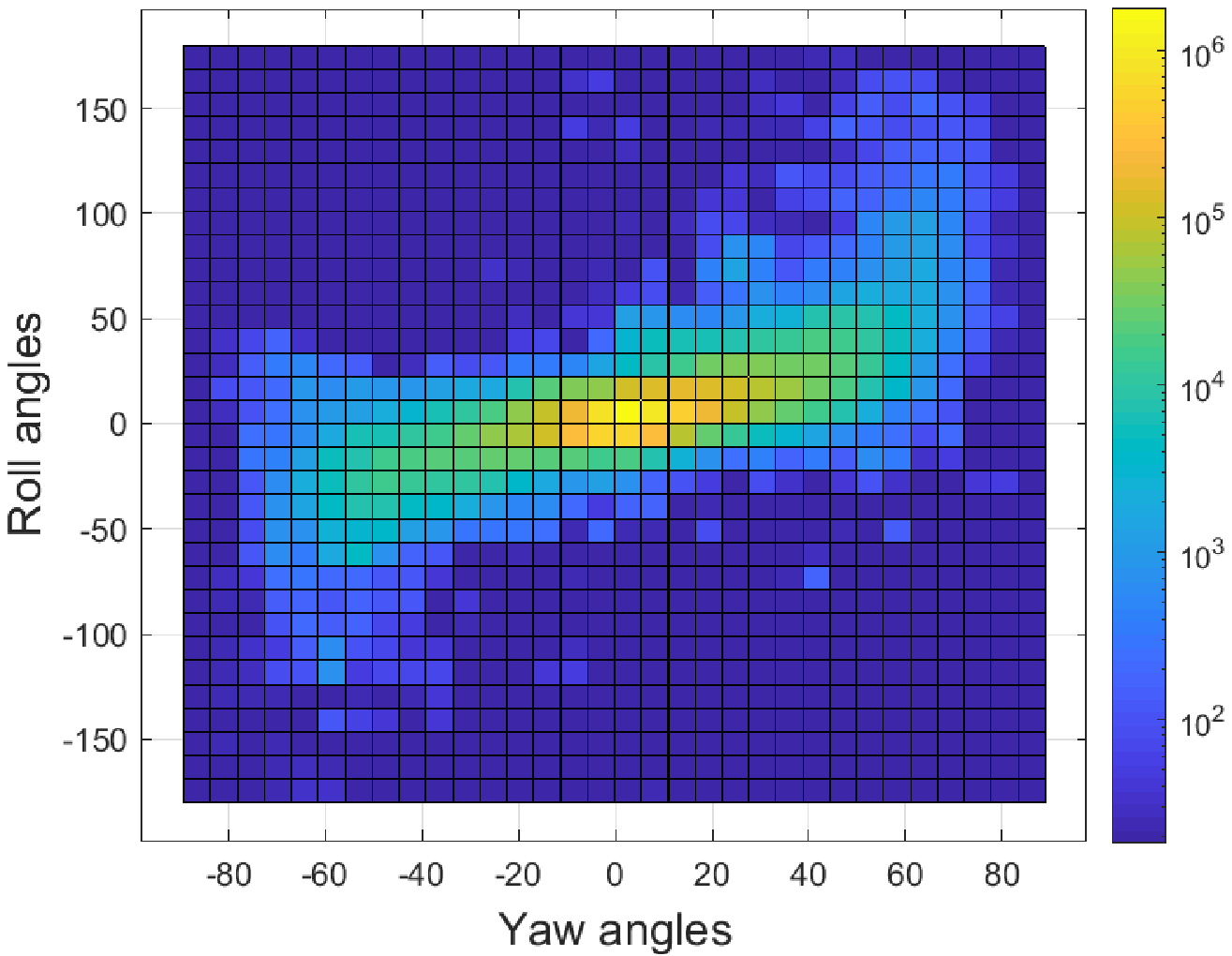}
\label{fig:yr}
}  \hspace{-0.5cm} 
\subfigure[Driver's head: roll-pitch angles]
{
\includegraphics[height=5.0cm]{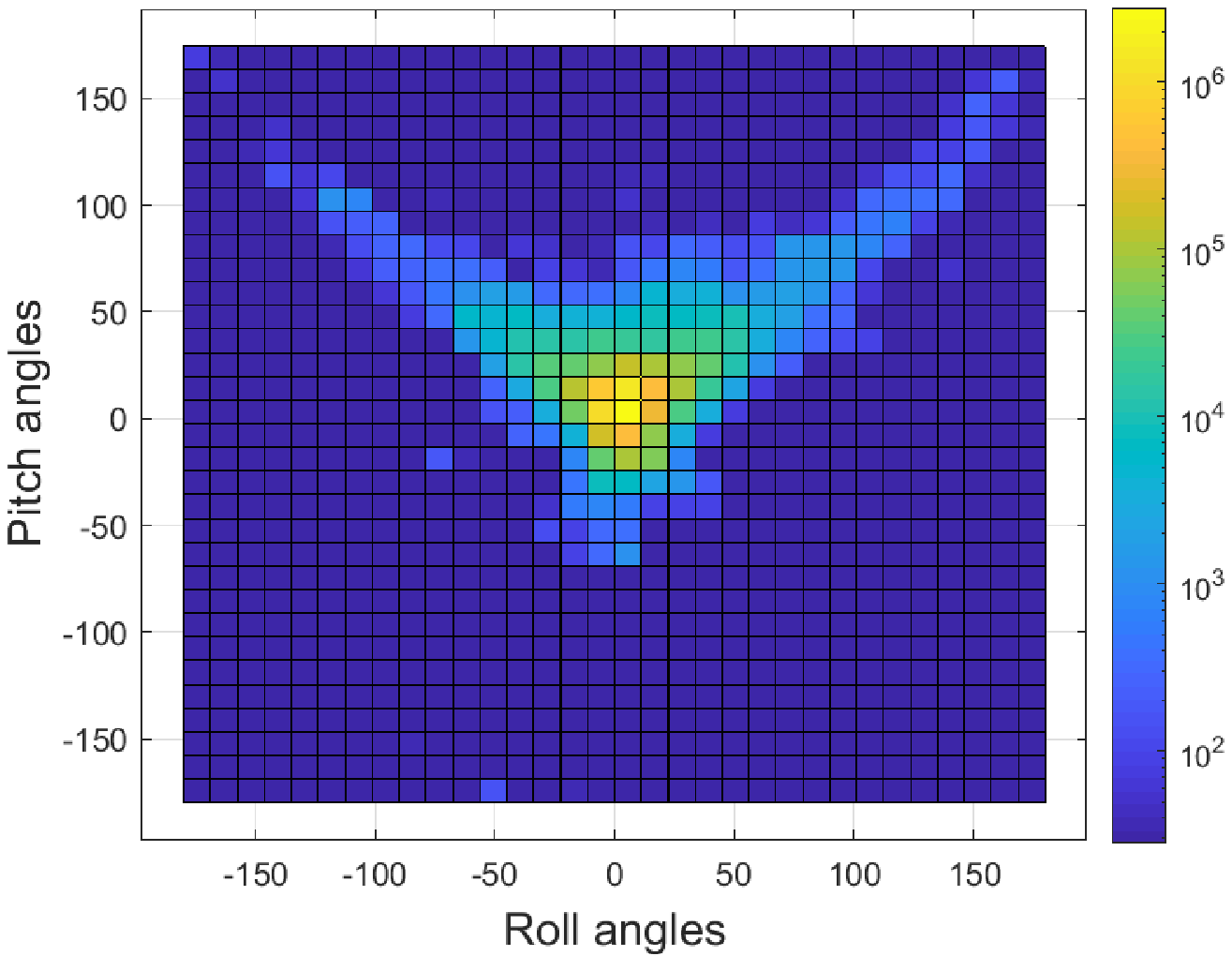}
\label{fig:rp}
}
\caption{Distribution of the driver's head orientation in our database estimated with the Fi-Cap helmet. The results are projected into the yaw-pitch, yaw-roll and roll-pitch spaces.} 
\label{fig:headPoseDist}
\end{figure*}

We start our analysis by showing the distribution of the driver's head orientation during the the entire recording. Figure \ref{fig:headPoseDist} shows the distribution, projected on the yaw-pitch, yaw-roll and roll-pitch spaces. We notice that our dataset covers a large range of head poses along all three rotation axes due to the large number of subjects included, and the variety of primary and secondary driving activities considered during the data acquisition. Figure \ref{fig:yp} shows a wide symmetric yaw angle range around the origin spanning between $\ang{-80}$ to $\ang{80}$, which reflects practical driving scenarios where the driver looks forward most of the time, but frequently checks the mirrors, dashboard, windshield and side windows. In contrast, pitch angles have an asymmetric range spanning from $\ang{-50}$ to $\ang{100}$, reflecting high degrees of freedom at high pitch angles as the driver looks at the dashboard or gear shifter compared to low pitch angles which almost vanishes at the edge between the windshield and the car ceiling.

\subsection{Driver's Gaze Distribution}
\label{ssec:DistributionGaze}

We also analyze the distribution of the annotated gaze directions. We represent the driver's gaze using the elevation (vertical) and azimuth (horizontal) gaze angles and report the distribution of these angles. First, we pick the frames where the driver is looking at target markers inside the car while driving  (Fig. \ref{Fig:marker}). The locations of these markers are estimated during the calibration phase for our reference. We approximate the head location with the location of the Fi-Cap helmet. With the head location, we define the driver's gaze as the line joining the head's location with the target marker and calculate the elevation and azimuth angles of this line in the 3D space. Hence, we can obtain absolute gaze directions that are agnostic of the camera locations and calibrations. Second, we take the frames where the driver is looking at the fiducial marker outside the car. The location of this marker is estimated with respect to the road camera. We transform the data into the back camera coordinate using the calibration data. The gaze direction is once again calculated in terms of the elevation and azimuth angles using the same method.

\begin{figure}[t]
\centering
\subfigure[Looking at in-vehicle markers]
{
\includegraphics[trim=0.3cm 0 2.6cm 0, clip, height=3.5cm]{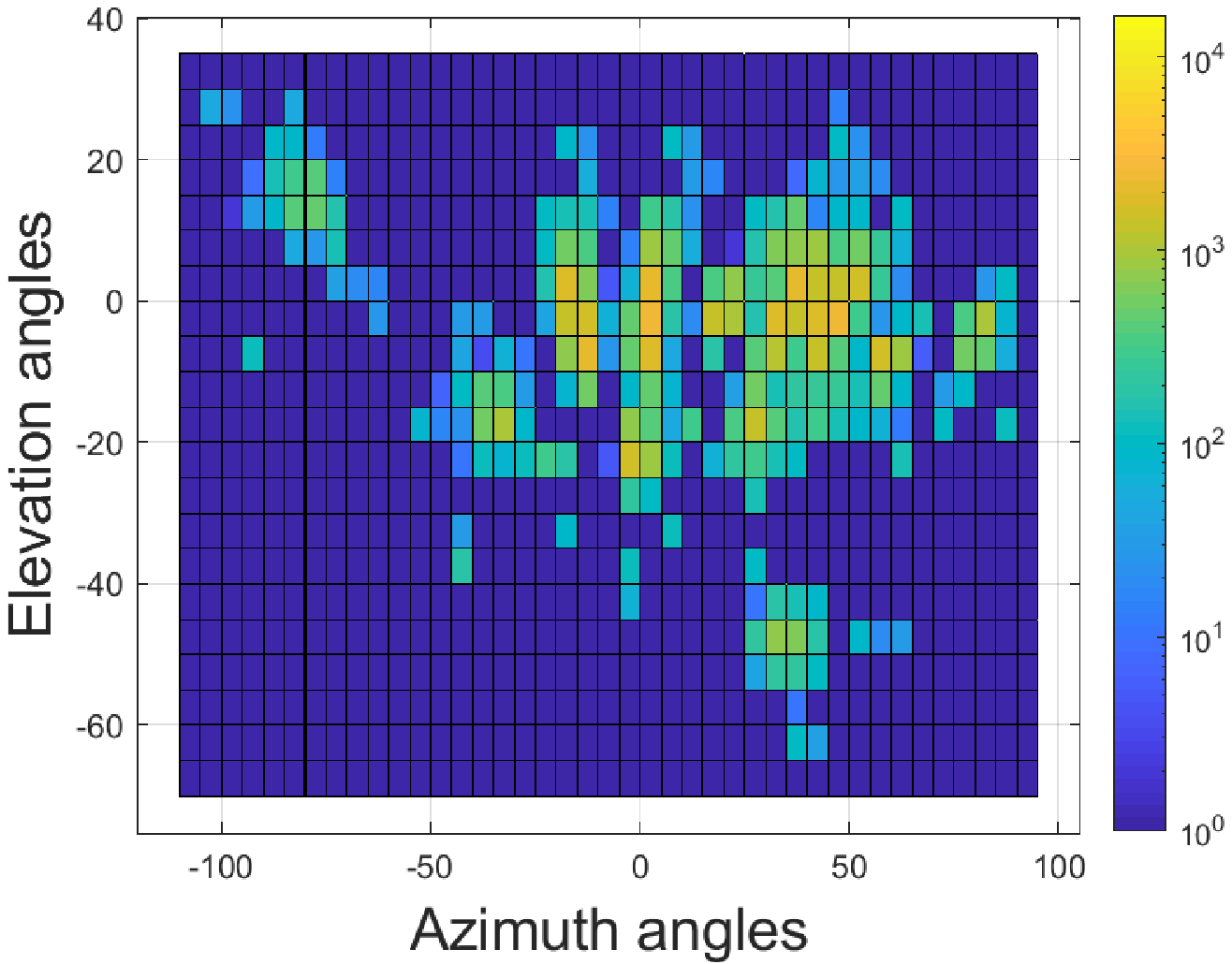}
\label{fig:discrete}
} \hspace{-0.4cm}
\subfigure[Looking at the moving board]
{
\includegraphics[height=3.5cm]{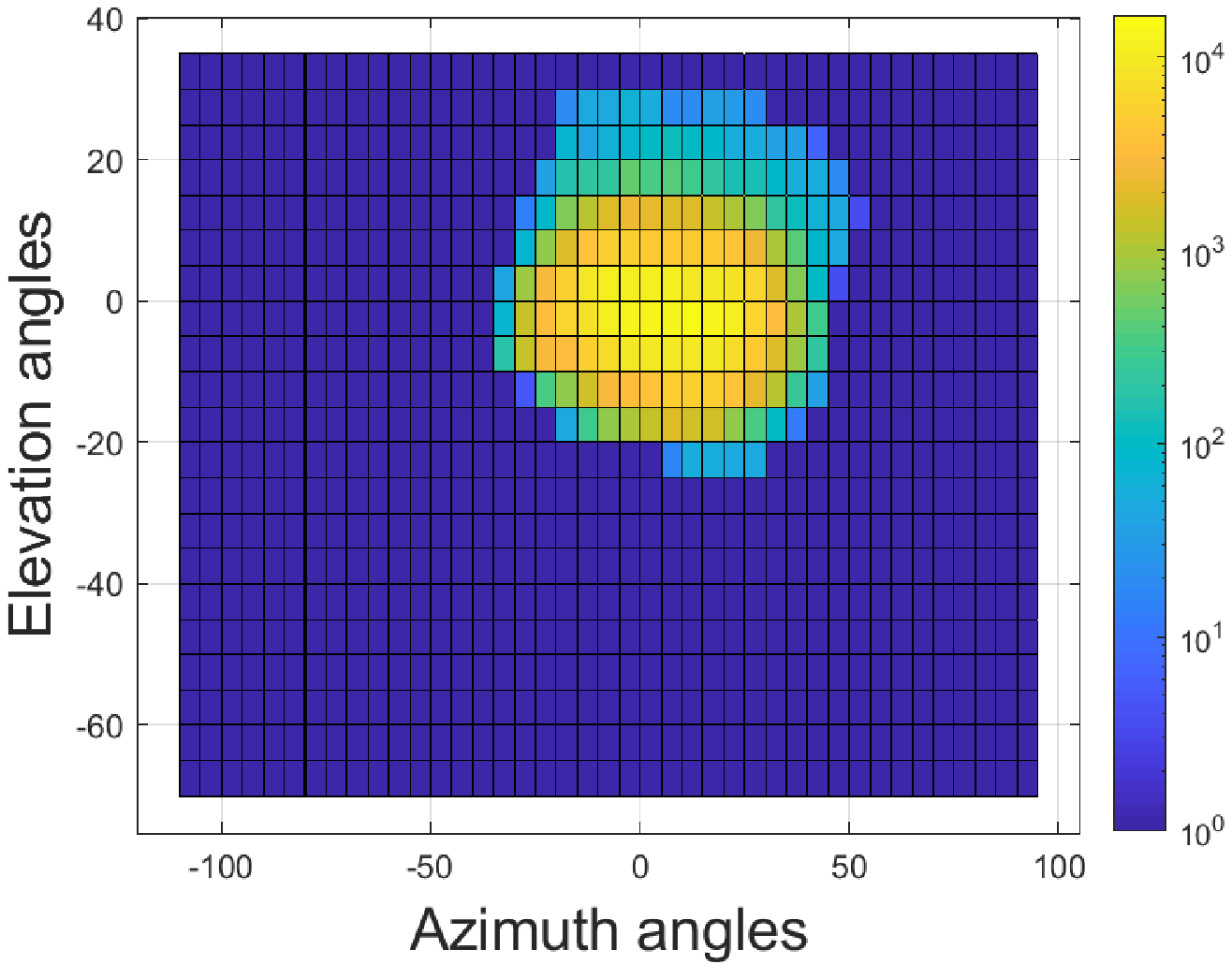}
\label{fig:combined}
} 
\caption{Distribution of the driver's gaze angles in our dataset: (a) distribution of the drivers' gaze as she/he looks at target markers inside the car (step 3 in our protocol), and (b) distribution of the driver's gaze angles as the driver looks at the moving board outside the car (step 1 in our protocol).} 
\label{fig:AnglesDist}
\end{figure}

Figure \ref{fig:discrete} reports the distribution of the gaze angles when the driver looks at markers inside the car while driving (step 3 in our protocol - Sec. \ref{ssec:protocol}). Figure \ref{fig:combined} reports the distribution of the gaze angles as the driver looks at the moving board during the parking phase (step 1 in our protocol - Sec. \ref{ssec:protocol}). As expected, the gaze angle distribution for the in-vehicle markers has a higher azimuth and elevation angles ranges since these markers span more locations in the car in contrast to the moving board gaze distribution, which only spans the windshield area. However, the  moving board gaze angles have a more balanced and smooth distribution due to their continuous nature. In addition, they can help in training time-based deep learning models (e.g., recurrent neural networks) as they provide continuous gaze annotations for successive frames.


The MDM database can be used for other gaze representations. For example, the driver's gaze can be represented in terms of normalized pixel location in the road camera scene. The 3D location of the target marker can be mapped into any camera as the reference frame, since all cameras are calibrated. This alternative representation can be useful if the goal is to map the driver's gaze directly to targets on the road.

\subsection{Head Pose Estimation with Depth Data}
\label{ssec:depth}

\begin{figure}[t]
\centering
\includegraphics[width=0.98\columnwidth]{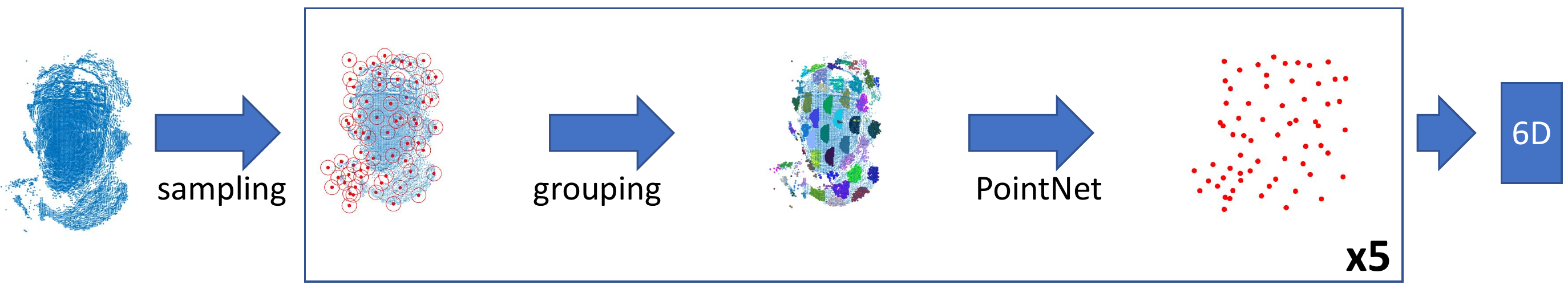}
\caption{Diagram of the algorithm proposed by Hu \etal \cite{Hu_2020} to estimate head pose of the driver using point cloud data. The framework directly uses the point cloud data without projecting the 3D points into 2D spaces.}
\label{fig:headpose_from_depth}
\end{figure}

This section demonstrates one of the potential uses of the MDM database for \emph{head pose estimation} (HPE). In Hu \etal \cite{Hu_2020}, we showed that we can effectively estimate head pose from a depth camera by directly processing point cloud data. In this work, we achieved better performance than state-of-the-art HPEs based on RGB images, especially in frames with large rotations. Figure \ref{fig:headpose_from_depth} shows the model structure inspired by the work of Qi \etal \cite{Qi_2017_2,Qi_2017}, where we formulate this task as a regression problem. We have three basic building blocks for this model: sampling, grouping and PointNet. In sampling, we use the iterative farthest point sampling algorithm to get ``anchor points.'' This step reduces the redundancy of the point cloud while maintaining its structure as much as possible. In the grouping building block, we group points within a radius $R$ of the anchor points, capturing the relationship between each anchor point and its neighbors. In PointNet, we adopt multi-layer perceptrons to learn from the data features that are more discriminative for our task. We repeat the sampling-grouping-PointNet layer set five times, aggregating features from different resolutions. Finally, we acquire a high-level feature vector that represents the input point cloud. We use a fully-connected layer to derive the head rotation from this feature vector. We retrain the models using data from 27 subjects for training, 10 for validation and 22 for testing. 

We estimate the \emph{mean square error} (MSE) of this model for each of the 22 subjects in the test set, reporting their average results in Table \ref{tab:MSE}. The average angular error for our estimation was between $\ang{5.91}$ and $\ang{8.79}$ on the entire test set. We compare our point cloud HPE model with a state-of-the-art HPE algorithm based on RGB images. We use OpenFace 2.0 \cite{Baltrusaitis_2017}, estimating the head pose with the 1,920 $\times$ 1,080 face camera (GoPro camera -- Fig. \ref{fig:fc}). To ensure that the definition of head pose in our dataset is consistent with OpenFace 2.0, for each subject, we find an average transformation between the head pose ground truth and OpenFace 2.0. We observe that OpenFace 2.0 fails at giving predictions in 5.69\% of the frames in the test set. The performances on the frames with predictions are $\ang{9.98}$ for roll, $\ang{9.13}$ for yaw and $\ang{9.60}$ for pitch angles. In contrast, the proposed approach produces a head pose estimate for every frame. For fair comparison, we evaluate our approach only on the frames with predictions from the RGB-based HPE algorithm, which we refer to as the \emph{OpenFace set}. Table \ref{tab:MSE} shows clear improvements in using our point cloud algorithm, with absolute differences of $\ang{4.28}$ for roll, $\ang{0.61}$ for yaw and $\ang{3.51}$ for pitch angles. 

\begin{table}[t]
\centering
\caption{MSE of the point-cloud based algorithm and OpenFace 2.0. We report results of the proposed model on the entire test set and also on the set where OpenFace 2.0 provides estimations. }
\vspace{-0.1cm}
\begin{tabular}{l| c c c | c}
\hline
Method && Errors&&Percentage Frame  \\ 
& Roll(\degree) & Yaw (\degree) & Pitch (\degree) & Missed\\ 
\hline\hline
Hu \etal &5.91	&8.79	&6.15 & 0\% \\ 
\hline
\multicolumn{1}{c}{}&\multicolumn{4}{c}{OpenFace set}\\
\hline
Hu \etal &\textbf{5.70} &\textbf{8.53}&	\textbf{6.08} & 0\% \\ 
OpenFace 2.0  &9.98 &9.13&	9.60 & 5.69\% \\
\hline
\end{tabular}
\label{tab:MSE}
\vspace{-0.4cm}
\end{table}

\section{Conclusions}
\label{sec:conclusion}

This paper presented the MDM database, which provides naturalistic recordings that are ideal for studying visual attention in realistic driving conditions. The database includes multimodal sensors including four HD cameras, deep sensors, grayscale camera with IR illumination, and a microphone array. The database also includes the CAN-Bus information to obtain vehicle related information. The database provides ground truth for both head pose and gaze information. The Fi-Cap helmet is used to provide continuous annotation of the driver's head pose. The placement of the helmet in the back of the driver's head avoids occlusions for frontal cameras recording the face of the subject. This setup provides head pose information for each of the millions of frames included in the corpus. We obtained the driver's reference gaze in a continuous setting, where we move a board with a fiducial point outside the car while the car is parked. We also collected the event based gaze directions by asking drivers to look at particular locations, both inside and outside the vehicle. In addition, we collected data when the driver performs natural secondary activities such as changing the radio station and following instructions. Collectively, the information included in the corpus can be useful in training efficient algorithms to study driver visual attention in naturalistic driving conditions. 

This study also provided an analysis of the MDM corpus. We introduced a novel experimental framework to quantify the robustness of our head pose labels by studying the estimated head pose as a function of the number of detected fiducial points in the Fi-Cap helmet. By analyzing the distribution of gaze and head pose information, we showed that our dataset covers a wide range of angles in all rotation axes, which reflects the variety of primary and secondary activities included in the data collection protocol. The collected data provides a great resource to analyze the driver behavior. Its size, the driver diversity, the naturalistic nature of the recording, the presence of multiple sensors and the detailed annotations provided in the corpus are ideal to train sophisticated deep learning solutions to monitor driver visual attention for in-vehicle systems.

\subsection*{Access to Corpus}

The multimodal driver monitoring dataset is licensed free of cost to academic institutions under a \emph{Federal Demonstration Partnership} (FDP) Data Transfer and Use Agreement. We have also established a license for commercial institutions interested in our corpus.


%

\appendices


\section*{Acknowledgment}
This work was supported by SRC / Texas Analog Center of Excellence (TxACE), under task 2810.014. The authors would like to thank Lucas Weaver, Mayank Mangla, Stanley Liu, Vijay Pothukuchi, and Muhammad Ikram from Texas Instruments and Aneesh Saripalli from UT-Dallas for helpful discussions on the subject of this paper.


\ifCLASSOPTIONcaptionsoff
  \newpage
\fi




\bibliographystyle{IEEEtran}
\bibliography{reference_new,reference}

%
%

\vspace{-0.4cm}

\begin{IEEEbiography}[{\includegraphics[width=1in,height=1.25in,clip,keepaspectratio]{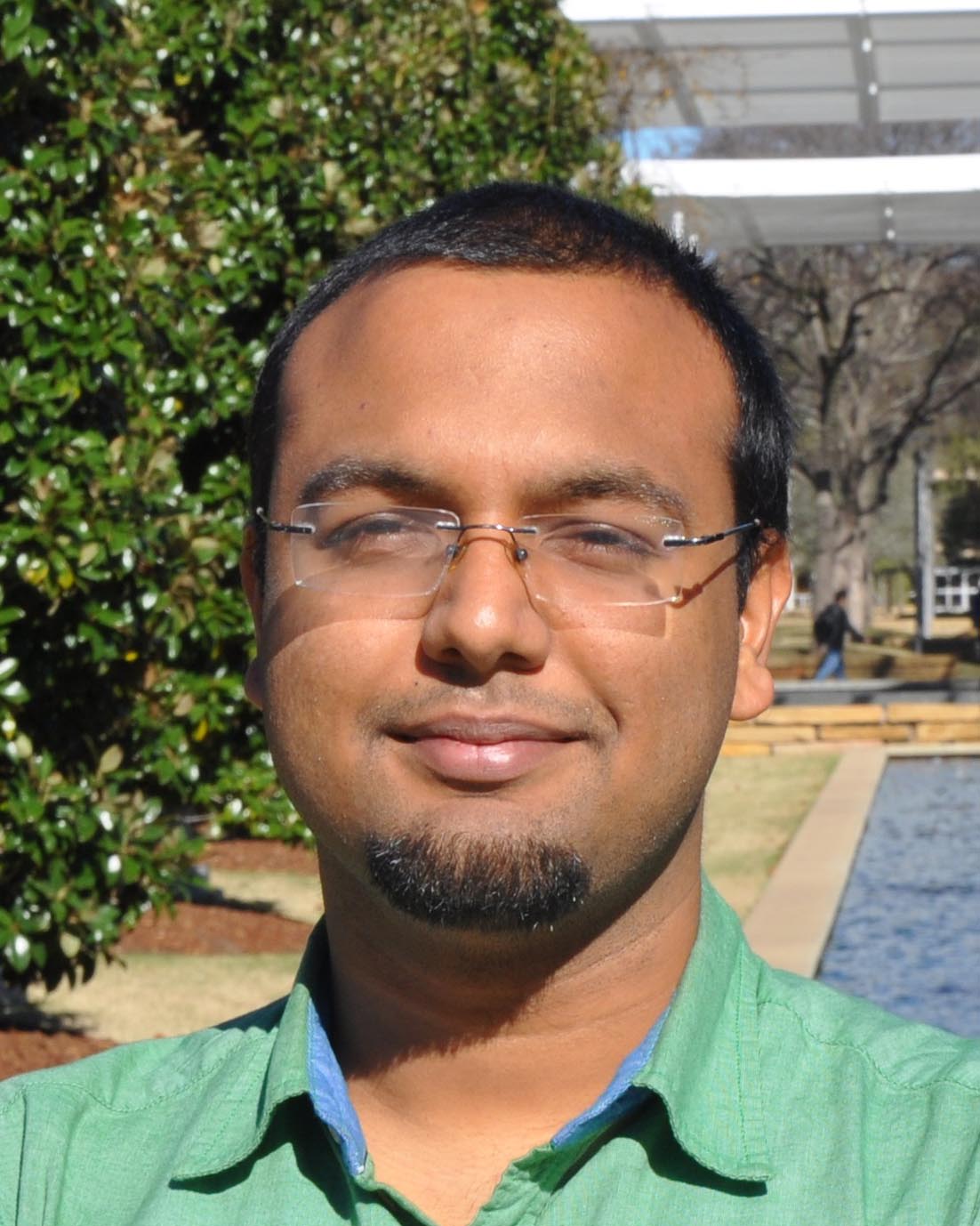}}]{Sumit Jha}
(S'16) received the B.Tech degree in Electronics and Communication Engineering form the National Institute of Technology(NIT), Trichy, India, in 2012 and the MS degree in Electrical Engineering from the University of Texas at Dallas (UTD), Texas in 2016. He is currently a PhD candidate at the University of Texas at Dallas. At UTD, he has been a part of the Multimodal Signal Processing(MSP) laboratory since 2015. His research interest lies in machine learning computer vision solutions for driver monitoring and in-vehicle safety systems. He has been a student member of IEEE since 2016.
\end{IEEEbiography}

\vspace{-0.4cm}

\begin{IEEEbiography}[{\includegraphics[width=1in,height=1.25in,clip,keepaspectratio]{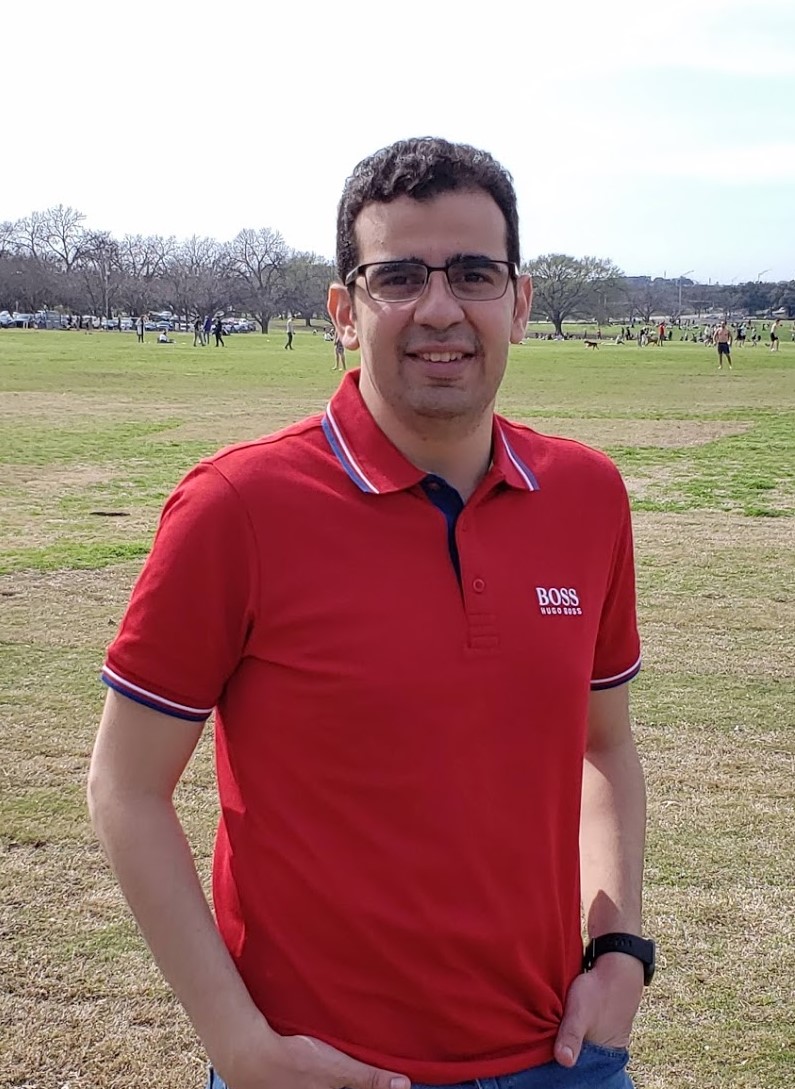}}]{Mohamed F. Marzban}
received the B.Sc. and M.Sc. degrees in electrical engineering from Cairo University, Egypt, in 2013 and 2016, respectively. He is currently pursuing the Ph.D. degree in electrical and computer engineering with The University of Texas at Dallas, Richardson, TX, USA. From 2013 to 2014, he was a Systems Research and
Development Engineer with Intel Labs, Cairo, Egypt. From 2015 to 2016, he was a Software Engineer with Avidbeam Inc., Cairo. In 2018, he joined Qualcomm Technologies., Santa Clara, CA, USA, as a connected vehicles Intern. He is the recipient of the Betty and Gifford Johnson Graduate Studies Award, Louis Beecherl Graduate Fellowship and the Jan Van der Zeil Graduate Fellowship in 2017, 2018 and 2019, respectively. His research interests include machine/deep learning applications,  advanced driver assistance systems and wireless communications. 
\end{IEEEbiography}


\vspace{-0.4cm}

\begin{IEEEbiography}[{\includegraphics[width=1in,height=1.25in,clip,keepaspectratio]{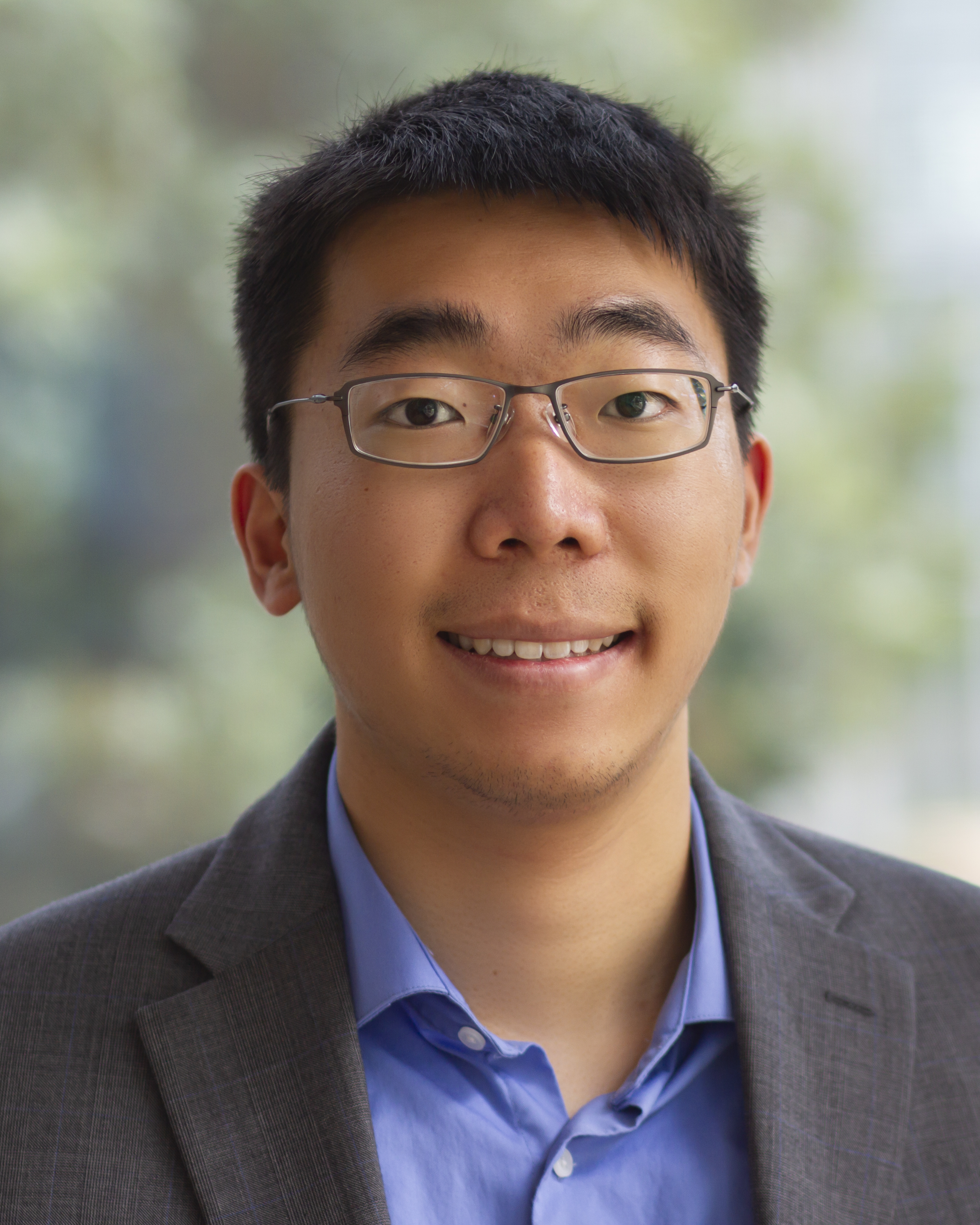}}]{Tiancheng Hu}
(S'20) received the B.Sc. in electrical engineering from The University of Texas at Dallas in 2020. During the 2018-2019 school year, He was the Assistant Director of Tutoring of the IEEE student branch at UTD. During his undergraduate studies, he also worked as an undergraduate researcher in the Multimodal Signal Processing (MSP) laboratory.  He was the recipient of the university-wide Undergraduate Research Scholarship as well as the Jonsson School
Undergraduate Research Award, in 2019 and 2020, respectively. His current research interests include computer vision and perception for advanced driver-assistance systems.
\end{IEEEbiography}

\vspace{-0.4cm}

\begin{IEEEbiography}[{\includegraphics[width=1in,height=1.25in,clip,keepaspectratio]{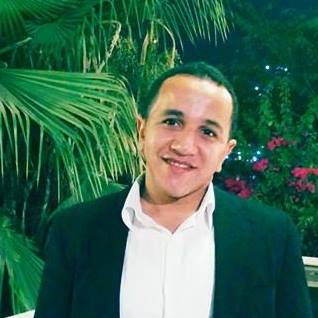}}]{Mohamed H. Mahmoud} received the B.Sc. and M.Sc. degrees in electrical engineering from Cairo University, Egypt, in 2007 and 2016, respectively. He is currently pursuing the Ph.D. degree in electrical engineering with The University of Texas at Dallas, Richardson, TX, USA. From 2010 to 2016, he was with Wasiela, Cairo, where he involved in the development of the LTE-UE and DVB. From 2016 to 2019, he joined RIOT-MICRO, Cairo, Egypt, as principle Systems Engineer, where he involved in the development of NB-IoT and LTE-M. His current research topics include machine/deep learning application,  advanced driver assistance systems and wireless communications. 
\end{IEEEbiography}

\vspace{-0.4cm}

\begin{IEEEbiography}[{\includegraphics[width=1in,height=1.25in,clip,keepaspectratio]{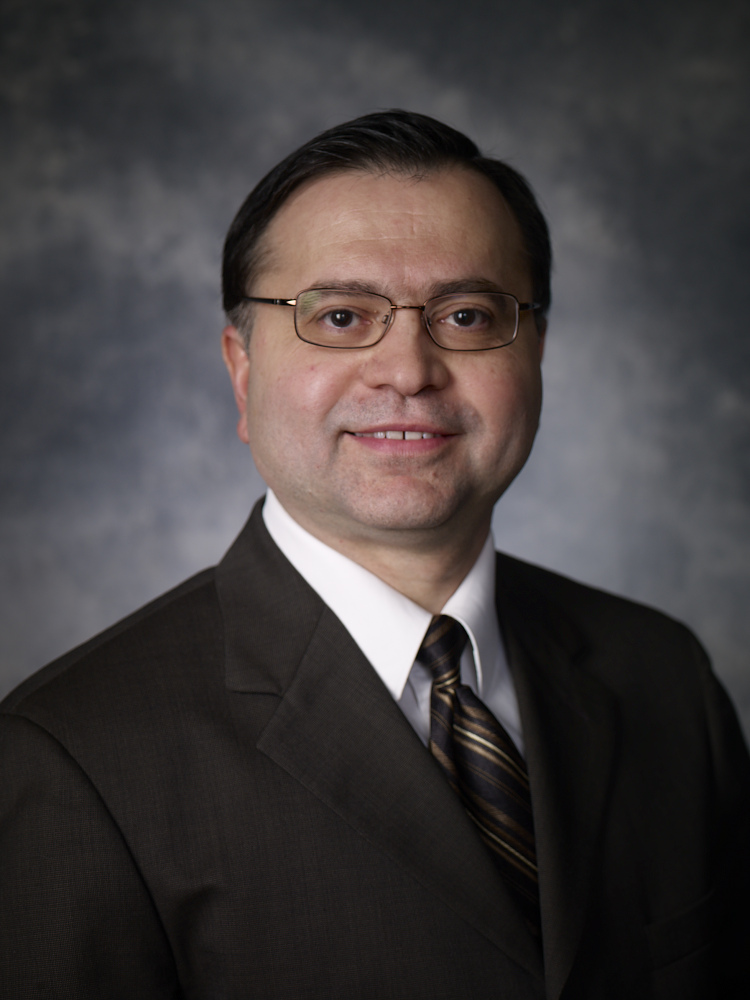}}]{Naofal Al-Dhahir} is Erik Jonsson Distinguished Professor $\&$ ECE Associate Head at UT-Dallas. He earned his PhD degree from Stanford University and was a principal member of technical staff at GE Research and AT$\&$T Shannon Laboratory.  He is co-inventor of 43 patents, co-author of 440 papers and co-recipient of 4 IEEE best paper awards. He is an IEEE Fellow, received 2019 IEEE SPCE technical recognition award, and served as Editor-in-Chief of IEEE Transactions on Communications.
\end{IEEEbiography}

\vspace{-0.4cm}

\begin{IEEEbiography}[{\includegraphics[width=1in,height=1.25in,clip,keepaspectratio]{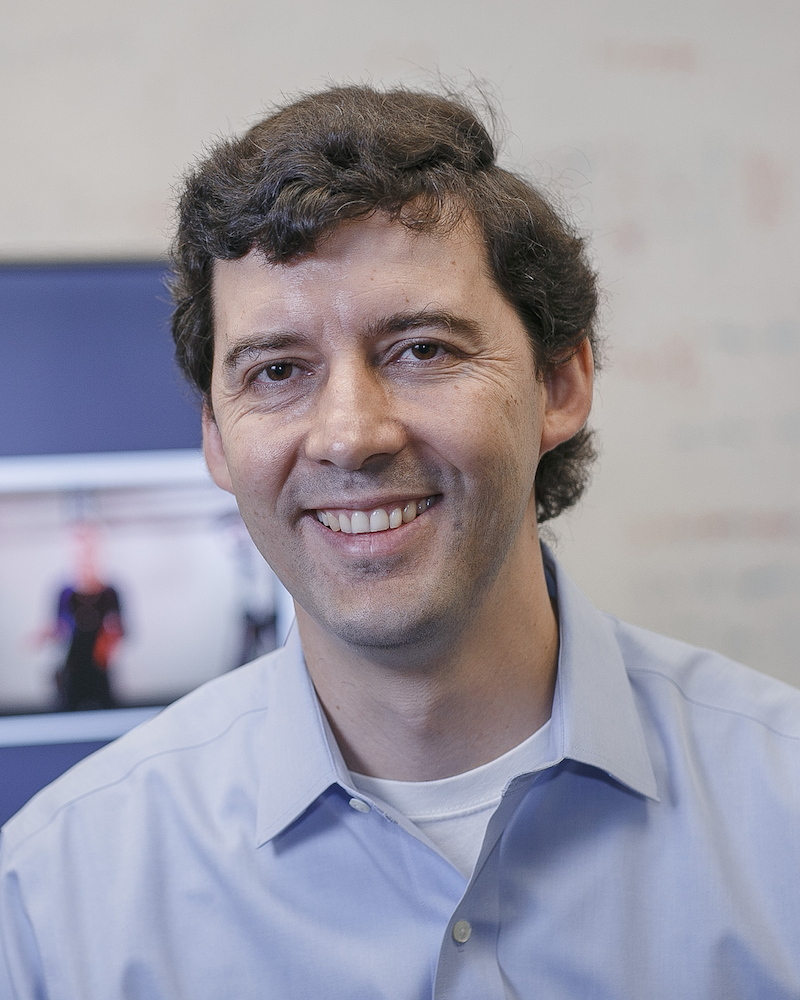}}]{Carlos Busso} 
(S'02-M'09-SM'13) received the BS and MS degrees with high honors in electrical engineering from the University of Chile, Santiago, Chile, in 2000 and 2003, respectively, and the PhD degree (2008) in electrical engineering from the University of Southern California (USC), Los Angeles, in 2008. He is a professor at the Electrical Engineering Department of The University of Texas at Dallas (UTD). He was selected by the School of Engineering of Chile as the best electrical engineer graduated in 2003 across Chilean universities. At USC, he received a provost doctoral fellowship from 2003 to 2005 and a fellowship in Digital Scholarship from 2007 to 2008. At UTD, he leads the Multimodal Signal Processing (MSP) laboratory [http://msp.utdallas.edu]. He is a recipient of an NSF CAREER Award. In 2014, he received the ICMI Ten-Year Technical Impact Award. In 2015, his student received the third prize IEEE ITSS Best Dissertation Award (N. Li). He also received the Hewlett Packard Best Paper Award at the IEEE ICME 2011 (with J. Jain), and the Best Paper Award at the AAAC ACII 2017 (with Yannakakis and Cowie). He is the co-author of the winner paper of the Classifier Sub-Challenge event at the Interspeech 2009 emotion challenge. His research interest is in human-centered multimodal machine intelligence and applications. His current research includes the broad areas of affective computing, multimodal human-machine interfaces, nonverbal behaviors for conversational agents, in-vehicle active safety system, and machine learning methods for multimodal processing. His work has direct implication in many practical domains, including national security, health care, entertainment, transportation systems, and education. He has served as general chair of ACII 2017 and ICMI 2021. He is a member of ISCA, AAAC, and a senior member of the IEEE and ACM.
\end{IEEEbiography}




\end{document}